%% file: elsarticle-template-num.tex
\documentclass[preprint, 3p, 12pt]{elsarticle}

\usepackage{amssymb}
\usepackage{booktabs}
\usepackage{amsmath,amsfonts}
\usepackage{amsthm}
\usepackage{multirow}
\usepackage{siunitx}
\usepackage{bm}
\usepackage[colorinlistoftodos,prependcaption,textsize=tiny]{todonotes}
\usepackage{lineno}
\newcommand{\blue}[1]{\textcolor{black}{#1}}

\journal{arXiv}

\begin{document}

\begin{frontmatter}


\title{Explainable AI Guided Unsupervised Fault Diagnostics for High-Voltage Circuit Breakers\tnoteref{label1}}
\author[1]{Chi-Ching Hsu\corref{cor1}}

\author[2]{Gaëtan Frusque}

\author[2]{Florent Forest}

\author[3]{Felipe Macedo}

\author[1]{Christian~M.~Franck}

\author[2]{Olga Fink}
% \ead{test@test.com}
\cortext[cor1]{Corresponding author. E-mail address: hsu@eeh.ee.ethz.ch}
\affiliation[1]{organization={High Voltage Laboratory, ETH Zurich},
            city={Zurich},
            country={Switzerland}}

\affiliation[2]{organization={Intelligent Maintenance and Operations Systems Laboratory, EPFL},
                city={Lausanne},
                country={Switzerland}}
                
\affiliation[3]{organization={Hitachi Energy},
                city={Zurich},
                country={Switzerland}}


% \title{}

% \affiliation[label1]{organization={},

% \affiliation[label2]{organization={},

% \author{}

% \affiliation{organization={},%Department and Organization

\begin{abstract}
Commercial high-voltage circuit breaker (CB) condition monitoring systems rely on directly observable physical parameters such as gas filling pressure with pre-defined thresholds. While these parameters are crucial, they only cover a small subset of malfunctioning mechanisms and usually can be monitored only if the CB is disconnected from the grid. To facilitate online condition monitoring while CBs remain connected, non-intrusive measurement techniques such as vibration or acoustic signals are necessary. Currently, CB condition monitoring studies using these signals typically utilize supervised methods for fault diagnostics, where ground-truth fault types are known due to artificially introduced faults in laboratory settings. This supervised approach is however not feasible in real-world applications, where fault labels are unavailable. In this work, we propose a novel unsupervised fault detection and segmentation framework for CBs based on vibration and acoustic signals. This framework can detect deviations from the healthy state. The explainable artificial intelligence (XAI) approach is applied to the detected faults for fault diagnostics. The specific contributions are: 1) we propose an integrated unsupervised fault detection and segmentation framework that is capable of detecting faults and clustering different faults with only healthy data required during training 2) we provide an unsupervised explainability-guided fault diagnostics approach using XAI to offer domain experts potential indications of the aged or faulty components, achieving fault diagnostics without the prerequisite of ground-truth fault labels. These contributions are validated using an experimental dataset from a high-voltage CB under healthy and artificially introduced fault conditions, contributing to more reliable CB system operation.
\end{abstract} 


% \begin{highlights}
% \end{highlights}


\end{frontmatter}


\input{introduction}
\input{related_work}
\input{methods}
\input{results}
\input{conclusions}

% \label{}





\bibliographystyle{IEEEtran}
\bibliography{ref}



% \end{thebibliography}
\end{document}

%% file: introduction.tex
% introduce the argument from the beginning that assets are aging and are often in good condition when they are scheduled to be taken out of operation. But since they critical for operation, we cannot do it without knowing their health condition and monitoring it over time.. Since we need to do retrofitting only non-intrusive monitoring devices are suitable. But we are not only interested in detecting the fault but also knowing the fault type… This is normally only possible with supervised learning approaches. Some segmentation approaches and residual analysis  have been  introduced…. However, the domain experts still need to identify the fault type and may struggle with it… XAI may help but usually only applicable in supervised setups.  We propose a way to provide them additional guidance through explainability in unsupervised setups.

\section{Introduction}
\label{sec:intro}
Circuit breakers (CB) are critical for ensuring safety and reliability in electrical transmission and distribution systems. They are designed to handle and interrupt both nominal and short-circuit currents and are usually not frequently switched, but they are often replaced after several decades of service to maintain reliable and safe functionality despite their infrequent operation. Therefore, many of these CBs may still be in good working condition when they are replaced. Delaying the replacement of CBs nearing their planned service life -- whether determined by regulatory guidelines or supplier recommendations -- can yield significant cost savings and environmental benefits, provided they continue to operate reliably and safely. Although CBs are designed and tested to be highly robust, and capable of withstanding severe operational stress, as with any other electro-mechanical system, their components are still subject to degradation over time, influenced by both operational and environmental factors, as with any electro-mechanical system~\cite{hu2022general}.

To ensure the reliable and safe operation of aging circuit breakers (CBs) and enable timely detection of deviations from normal operation, it is crucial to implement a condition monitoring system. Such a system typically involves a data acquisition setup, which utilizes various types of sensors such as current sensors, and a data analysis algorithm that evaluates the collected signals to assess the health condition of the CBs. By using such monitoring systems, any deviations from the healthy condition can be detected promptly, allowing for the repair or replacement of CBs before they fail. 

Many condition monitoring parameters have been studied to assess the condition of various mechanical and electrical CB components, such as springs, dampers, latches, coils, contacts, and motors~\cite{razi2020condition}. Any of these components, individually or in combination, can be sources of faults that may lead to severe consequences. In this work, faults refer not to power system faults that CBs need to clear, such as terminal or short-line faults, but to faults in CB components themselves, such as spring or damper faults. In recent years, researchers have used different parameters for evaluating the CB condition. Commonly used parameters include coil current~\cite{razi2016circuit, pan2019approach}, travel curves~\cite{razi2015applicability, rudsari2019fault}, dynamic contact resistance~\cite{abdollah2018intelligent, liu2018prediction}, operation timing~\cite{razi2015applicability, rusek2008timings}, acoustic emissions~\cite{sugimoto2019study, iwata2022development, darnsomboon2022field}, and vibration~\cite{ye2022novel, hoidalen2005continuous, qi2020mechanical}. \blue{In particular, vibration signals and acoustic emissions have gained increasing attention for their non-intrusive, real-time monitoring capabilities~\cite{razi2020condition, tan2023review, zhou2023systematic}. Since vibration and acoustic sensors can be installed without affecting the integrity or functionality of the CB, they enable continuous condition monitoring without the need to disconnect the CB from the grid.  This makes them a superior alternative to traditional methods such as dynamic contact resistance measurement, which are often intrusive and impractical for real-time or long-term monitoring.} 

Based on these parameters, the condition of CBs can be monitored, allowing for condition assessment over time and fault detection. Algorithms applied to fault detection aim to train a model that learns the healthy sample distribution. Any deviation from this healthy distribution is considered as a potential fault. Fault detection has been performed in various fields such as turbofan jet engines~\cite{chao2019hybrid, nejjar2024domain}, wind turbines~\cite{marugan2019reliability, jimenez2019dirt, frusque2024non}, and CBs~\cite{obarcanin2023condition}. While these fault detection approaches have demonstrated success in detecting faults across various fields using only healthy data, they usually do not provide additional information regarding the specific fault types.

% Commonly used fault detection approaches include reconstruction-based methods~\cite{kingma2013auto, bergmann2018improving}, which try to reconstruct the outputs with model trained on healthy data and measure the discrepancy between the measured values and the predicted outputs, and one-class classification-based (OCC-based) methods~\cite{scholkopf2001estimating}, which try to learn a boundary of healthy data. 

In addition to only detecting the faults, it is important to distinguish between different fault types, with or without explicitly labeling them. For example, fault segmentation involves grouping faulty samples using unsupervised clustering methods, but fault segmentation alone does not inherently provide information regarding the specific fault types. Various fault segmentation approaches have been proposed for different systems~\cite{chao2021implicit, hsu2023comparison}, but such approaches have not yet been applied to CBs. Furthermore, while they can identify different clusters, determining which cluster corresponds to a particular fault type typically requires domain knowledge. In straightforward cases, where a fault type is linked to deviations in a small subset of features, this task may be relatively simple. However, for more complex fault patterns, experts may struggle to assign fault type labels and may require additional guidance.

% An alternative approach is to analyze the patterns obtained from the fault detection algorithms to differentiate between the different fault types.  Furthermore, fault segmentation alone does not inherently provide information regarding the specific fault types. 

Contrary to fault segmentation, fault diagnostics goes one step further and aims to identify specific fault types, which is typically achievable only through supervised learning approaches where labels are available. In current CB condition monitoring research, fault diagnostics is generally performed by training a supervised model. Existing CB works predominantly focus on fault diagnostics with artificially introduced fault conditions such as mechanism jam and spring shedding~\cite{zhao2019fault} and loose fixing bolt, electromagnet jamming, buffer failure, and high operating voltage~\cite{ye2022novel}, providing ground-truth labels. The objective of all these methods is to demonstrate that the models can differentiate between healthy conditions and various known fault types. However, obtaining labels for CBs in real operations, without having a training dataset with artificially induced faults, is challenging. In addition, it is difficult, if not impossible, to collect data representing every possible fault type~\cite{zio2022prognostics, floreale_sensitivity_2024, hoffmann2020integration}. More details about fault detection, segmentation, and diagnostics are summarized in Figure~\ref{fig:faults} and in Section~\ref{sec:fault_d_s_d}.

% \blue{These signals are non-stationary and have a short duration, lasting only a few hundred milliseconds. }

In this work, we propose an unsupervised fault detection and segmentation framework enhanced with an eXplainable Artificial Intelligence (XAI) guided fault diagnostics approach to improve the reliability of CB systems. First, faulty samples are detected using an autoencoder (AE). Subsequently, these faulty samples are clustered into different groups, separate from the healthy cluster, indicating potential faulty conditions but without providing explanations for the faults. To provide insights into potential fault types and support domain experts in diagnosing these conditions, we incorporate an XAI approach to explain the faults. Typically, XAI approaches explain the rationale behind the model outputs and are usually applicable only in supervised setups where labels are available. Since no labels are available in our case, we propose integrating a classifier into the AE that represents the cluster separation achieved through clustering. This integration makes supervised XAI methods applicable to explain the clusters identified in our unsupervised fault segmentation framework. Previous XAI approaches have mainly focused on the computer vision domain and, to the best of our knowledge, have not been applied to CB condition monitoring data in an unsupervised way. The proposed framework is evaluated using an experimental dataset from a high-voltage CB, with non-intrusive measurements including vibration and acoustic signals recorded during open operations under healthy and artificially introduced fault conditions. Finally, we show the flexibility of this framework by conducting experiments using various clustering methods ($K$-means, OPTICS, and Self-Organizing Maps) offline and online, and quantitatively assess the quality of the resulting explanations using an XAI approach, Integrated Gradients.

% summarise
The main contributions of the present work are summarized as follows:
\begin{enumerate}
    % \item \blue{We propose an integrated and flexible unsupervised fault detection and segmentation framework for CBs, capable of detecting and clustering different faults using only non-intrusively collected healthy data samples for training.}
    \item \blue{We design an unsupervised XAI-guided fault diagnostics approach, which integrates XAI techniques to provide explanations for the assignment of a sample to a specific cluster obtained in the fault segmentation process, even in the absence of ground-truth fault labels.}
    \item \blue{We apply our framework to experimental CB data collected in the laboratory for four different fault types, demonstrate its effectiveness and flexibility using various clustering methods offline and online, and assess the quality of the resulting explanations.}
\end{enumerate}

The remainder of the paper is structured as follows. The relevant literature on fault detection, segmentation, diagnostics, CB fault diagnostics in particular, as well as XAI, is provided in Section~\ref{sec:related_work}, while Section~\ref{sec:methods} introduces the fault detection, segmentation, and XAI methods. Section~\ref{sec:experiment} details the case study, including experimental setup for data collection. Section~\ref{sec:results} presents the results from fault detection, segmentation, and the results obtained using XAI on the experimental dataset, and the influence study on using different combinations of sensors in different directions and microphone. Section~\ref{sec:conclusions} presents the conclusions and the future research possibilities. 

%% file: related_work.tex
\section{Related work}
\label{sec:related_work}
% \subsection{Condition monitoring with non-intrusive measurement signals on CB}
% For instance, the work~\cite{yang2019chaotic} demonstrates that features extracted from vibration signals change when a fault presents during both opening and closing operations. 

% In literature, a CB is monitored with two microphones installed in a semi-anechoic chamber~\cite{sugimoto2019study}. The results indicate that it is possible to distinguish between three artificially introduced faults and the healthy condition based on frequency-domain features, but the background noise level is lower in an anechoic chamber than in substation and therefore the transferability is challenging. Another study conducted on a spring-operated high-voltage CB similarly demonstrates the ability to differentiate between various artificially introduced faults~\cite{iwata2022development}. 

\subsection{Fault detection, segmentation, and diagnostics}
\label{sec:fault_d_s_d}
The first step in condition monitoring, following data collection, is typically fault detection. This task involves identifying data samples with irregular distributions that deviate from the healthy data distribution~\cite{pang2021deep}, indicating a potential fault condition. Commonly used approaches include reconstruction-based methods~\cite{kingma2013auto, zhang2022anomaly, yang2022method}, one-class classification-based methods~\cite{scholkopf2001estimating, tax2004support}, and knowledge distillation (also called teacher-student framework)~\cite{bergmann2020uninformed, salehi2021multiresolution, deng2022anomaly}. Reconstruction-based methods, such as Autoencoders (AE), are trained on healthy data to learn the healthy data distribution. When anomalous data are input, these models exhibit higher reconstruction errors compared to healthy data~\cite{ruff2021unifying}. For example, AEs with different loss functions are used to detect faults in images in an unsupervised way~\cite{bergmann2018improving, bergmann2019mvtec}, while the anomalous regions can be segmented automatically. 

% Commonly used fault detection approaches include reconstruction-based methods~\cite{kingma2013auto, bergmann2018improving}, which try to reconstruct the outputs with model trained on healthy data and measure the discrepancy between the measured values and the predicted outputs, and one-class classification-based (OCC-based) methods~\cite{scholkopf2001estimating}, which try to learn a boundary of healthy data. 

Once a fault is detected, the faulty samples should be further investigated and diagnosed. Traditional fault diagnostics problems are usually formulated within a supervised learning framework, where labels are available~\cite{li2024novel, xu2022attention}. However, in real-world applications, including condition monitoring of CBs, an unknown number of faults could occur and ground-truth labels are unavailable or incomplete. Therefore, fault diagnostics tasks can be reformulated into fault segmentation tasks in an unsupervised way due to the lack of labels~\cite{floreale_sensitivity_2024}. Fault segmentation focuses on discriminating various fault types without identifying the cause of faults, e.g., which component is faulty. For instance, in turbofan jet engines, faults can be detected and segmented based on sensor-wise residuals from a reconstruction-based method~\cite{hsu2023comparison}. Further fault diagnostics can be achieved by analyzing the patterns of these sensor-wise residuals. 

\begin{figure*}[htbp]
\centerline{\includegraphics[trim={0.5cm 3.0cm 0.5cm 0.8cm}, width=\textwidth]{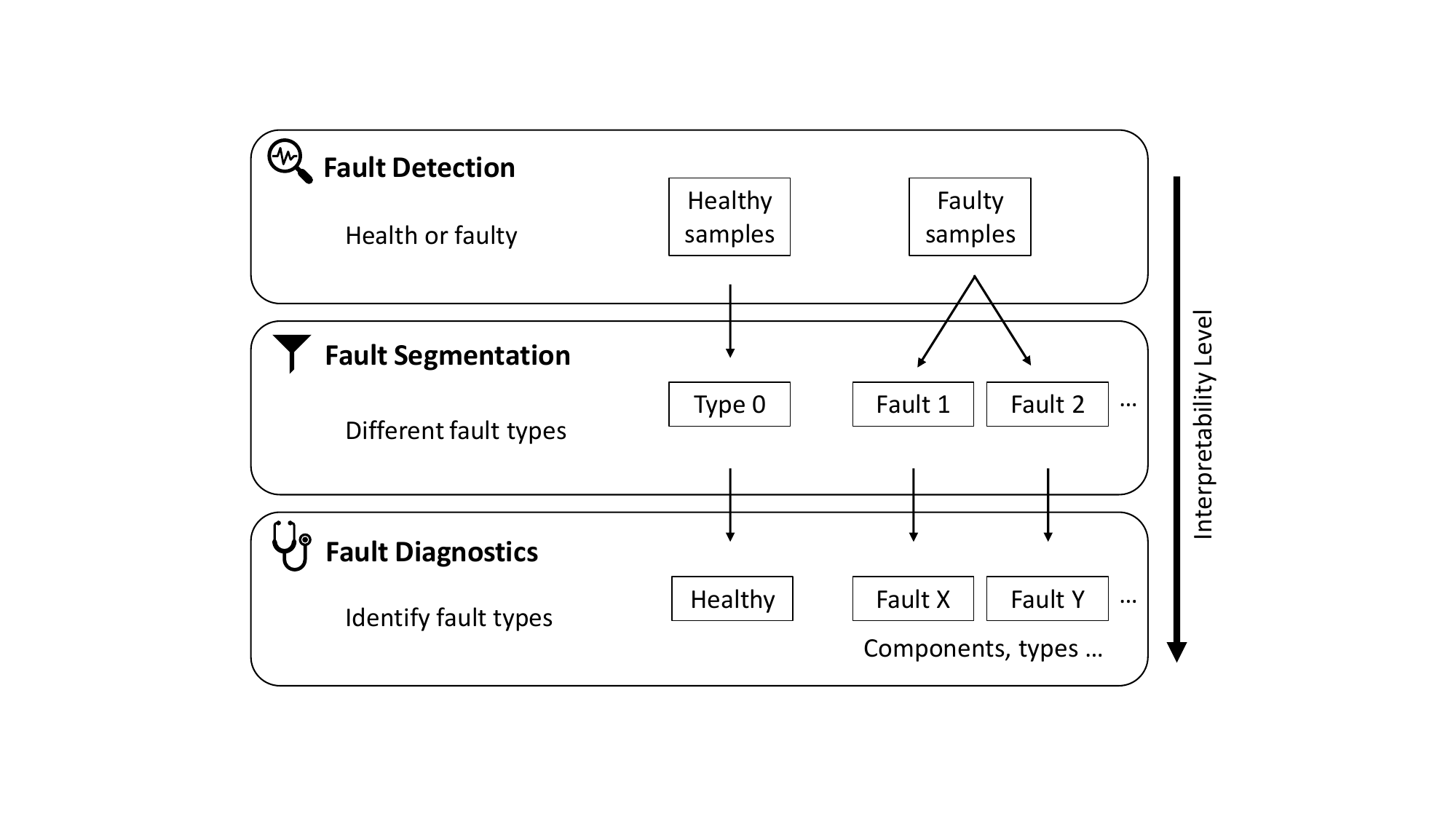}}
\caption{Scope of fault detection, segmentation, and diagnostics, modified from~\cite{chao2021implicit}. Fault detection aims to identify faulty samples from healthy samples. Fault segmentation differentiates between various fault types without diagnosing them. Fault diagnostics goes a step further by identifying the specific fault types such as identifying faulty components. Condition monitoring data can serve as input to any of these stages independently, depending on the application.}
\label{fig:faults}
\end{figure*}

The distinctions between fault detection, segmentation, and diagnostics are summarized in Figure~\ref{fig:faults}. The interpretability level increases through these stages. Fault detection identifies samples that deviate from the healthy data distribution, fault segmentation differentiates between various fault types, and fault diagnostics provides details about specific fault causes or components involved. Each stage adds additional information and explanation about potential fault types. Notably, these processes are not sequential; condition monitoring data can serve as input to any of these stages independently, depending on the application.

% OCC-based approaches, such as one-class support vector machines (OC-SVM)~\cite{scholkopf2001estimating} and support vector data description (SVDD)~\cite{tax2004support}, learn a compact one-class distribution that represents the boundary of the system's health condition~\cite{michau2020feature}. This distribution can then be used to detect faults or distinguish between different severity levels of fault conditions~\cite{michau2017deep}. 

% Another direction for AD is based on knowledge distillation (also the called Teacher-Student model)~\cite{bergmann2020uninformed, salehi2021multiresolution, deng2022anomaly}, where the pre-trained teacher model is used to learn the healthy distribution. The anomalies are detected when the outputs of the student model differ from the teacher model.

% In this work, we focus on clustering of the detected anomalies, assuming that similar fault types belong to the same cluster. In addition, these clusters can then be explained or inspected using the XAI methods, providing insights into the potential fault types or components.

% The third approach, knowledge distillation (also called Teacher-Student model),~\cite{bergmann2020uninformed}.
%  trim={<left> <lower> <right> <upper>}

\subsection{Circuit breaker fault diagnostics}

Existing literature on condition monitoring for CBs primarily focuses on fault diagnostics using supervised learning, often relying on artificially introduced faults. These studies can be categorized into three main directions: signal analysis, machine learning (ML), and deep learning (DL)-based~\cite{tan2023review}. First, the signal analysis method extracts statistical features from time-domain~\cite{qi2020mechanical}, frequency-domain, or time-frequency domain. Then, the extracted features are used to distinguish between different faulty conditions. Second, the ML-based methods utilize similar extracted features, but train classifiers to discriminate between the faulty samples. For instance, a one-class support vector machine (OCSVM) classifier is used with Wavelet transform features to first detect faults and then a supervised SVM classifier is trained to distinguish jam fault of the iron core, base screw looseness, and lack of lubrication~\cite{huang2015mechanical}.

DL-based methods have the advantage of the absence of a feature extraction step, where the DL algorithm is able to learn features by itself during training. For example, vibration signals are transformed into 3D time-frequency images by Hilbert-Huang transform and a 2D-CNN model is used to assess seven different damper conditions~\cite{yang2019condition}, or they are transformed into 2D time-frequency spectrograms by continuous wavelet transform (CWT) and a deep convolutional generative adversarial network (DCGAN) is used for data augmentation, increasing the amount of faulty samples, and a 2D-CNN model based on 2D time-frequency spectrograms is used to distinguish between different fault types~\cite{yang2024small}. A U-Net with CapsNet is proposed to identify five different fault types~\cite{ye2022novel}. Attention mechanisms and few-shot transfer learning techniques are employed for CB fault diagnostics to overcome the data scarcity challenge, which arises due to CBs' infrequent operations~\cite{wang2022few}.

\subsection{eXplainable Artificial Intelligence (XAI)}

The realm of XAI addresses the well-known challenge of black-box models in machine learning, aiming to explain and understand the rationale behind model predictions, making their decision process more transparent and fostering user trust~\cite{arrieta_explainable_2019}. This is particularly crucial in high-stakes engineering applications including CBs with potential impacts on safety, availability, and costs~\cite{pashami_explainable_2023,cummins_explainable_2024}. When a model is not inherently interpretable, such as linear models or small decision trees, post-hoc explainability is employed. This involves generating explanations for an already trained back-box model. 

A common post-hoc XAI method is feature attribution, which assigns an importance score to each feature to explain the model's prediction. Feature attribution method can be divided into three main categories: occlusion-based, gradient-based, and propagation-based~\cite{samek_explaining_2021}. Occlusion-based (or perturbation-based) methods measure how the prediction will change if certain features are missing or corrupted~\cite{zeiler2014visualizing}. Gradient-based methods~\cite{baehrens10a, sundararajan2017axiomatic} compute the model's gradients at a given input sample with respect to each input feature. A large gradient indicates that the input feature is important for the output prediction. Lastly, propagation-based methods~\cite{montavon2019layer, bach2015pixel} utilize propagation rules similar to the backward propagation used during neural network training to propagate the output predictions back to the input features. Attribution methods such as Class Activation Maps (CAM), Grad-CAM, Integrated Gradients, Shapley Additive explanations (SHAP), LIME and Layer-wise Relevance Propagation (LRP) have been applied to fault diagnosis based on time-domain~ \cite{li_explainable_2021,zhu_decoupled_2021,chen_explainable_2023,lomazzi_explainability_2023}, frequency-domain~\cite{chen_vibration_2020,kim_explainable_2021,decker_does_2023} and time-frequency domain signals~\cite{grezmak_explainable_2019,han_study_2022,sanakkayala_explainable_2022}. In the field of CB, the SHAP method has been applied to a CNN model with time-frequency spectrograms extracted from the vibration signals as inputs to explain the artificially introduced faults, in a supervised learning setting~\cite{tan2024fault}.

% unsupervised 
While these methods are overwhelmingly applied in supervised classification or regression settings, attributions can also be obtained in the cases of anomaly detection \cite{montavon2020explaining,gama_fault_2024} and clustering, using a classifier mimicking the clustering's decision boundary \cite{kauffmann2022clustering}. The present work focuses on the unsupervised setting, which has not been explored for CB applications.

% other types of explanations
Finally, there are multiple other ways to explain machine learning models. While attribution methods rely on the low-level input features, decisions can be explained by higher-level concepts~\cite{koh_concept_2020,forest_interpretable_2024}. Other types of explanation are case-based reasoning and explanation by examples, sometimes called prototypes~\cite{li_deep_2017,chen_this_2019}, and counterfactual explanations~\cite{verma_counterfactual_2022}, \textit{i.e.}, finding a small change that would lead to a different model outcome. For instance, the work~\cite{barraza2024scf} utilizes counterfactual explanations for interpretable fault diagnosis.

%% file: methods.tex
\section{Methods}
\label{sec:methods}
In this section, we introduce the proposed framework, which consists of three steps shown in Figure~\ref{fig:framework_ae}. The first step is to train a convolutional autoencoder (CAE) for fault detection, which learns the healthy data distribution. In the second step, fault segmentation, samples are grouped into various clusters, and the pseudo-labels are created. In the third step, an additional classifier is trained with the pseudo-labels obtained from the fault segmentation step for XAI-guided fault diagnostics to provide explanations for the segmented faults, supporting domain experts in diagnosing different faults.

% \begin{figure*}[htbp]
% \centerline{\includegraphics[width=\textwidth]{figures/framework.pdf}}
% \caption{Visualized summary of the proposed framework.}
% \label{fig:framework}
% \end{figure*}

\subsection{Convolutional Autoencoder (CAE)}
\begin{figure*}[htbp]
\centerline{\includegraphics[width=\textwidth]{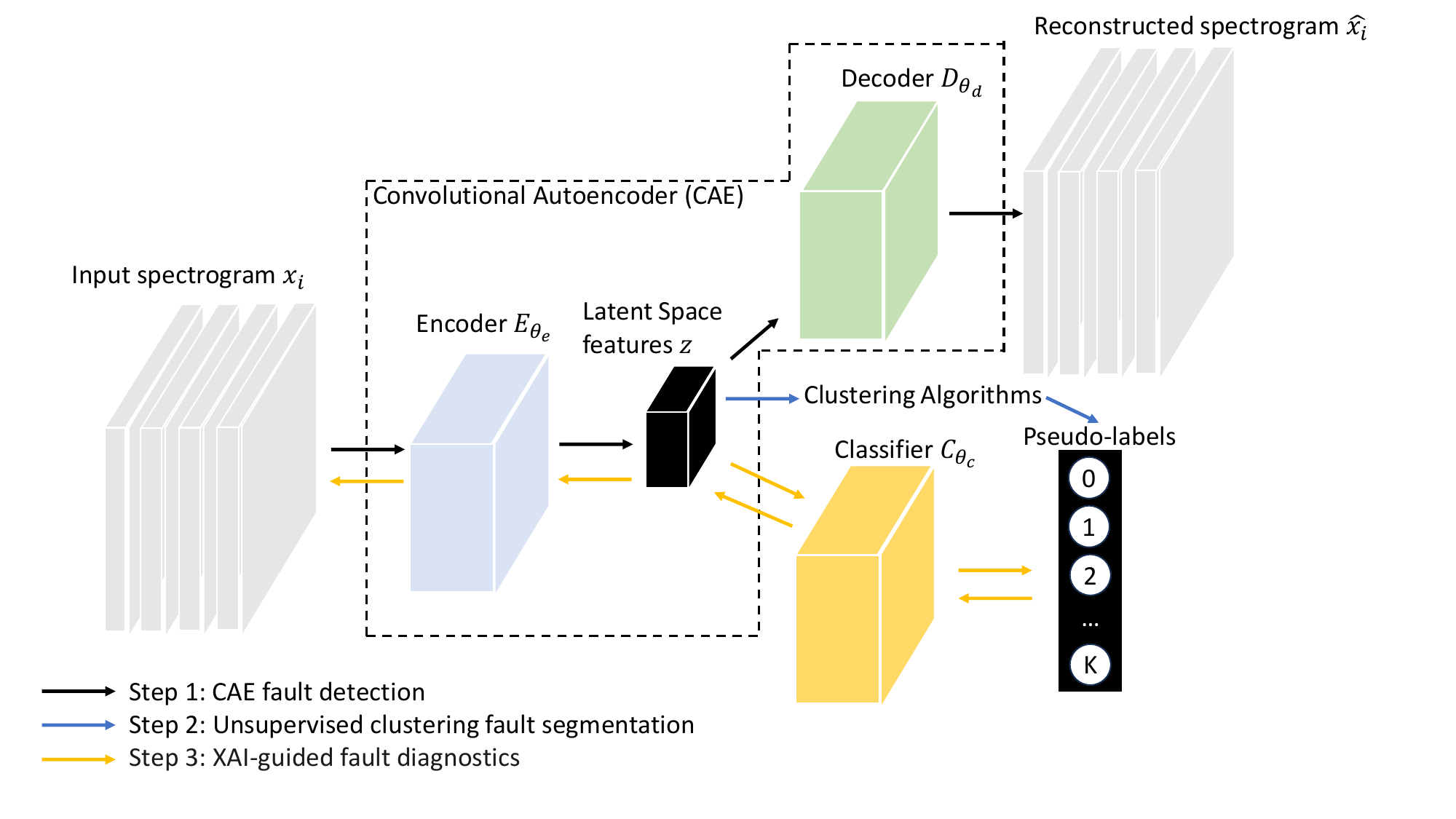}}
\caption{Overall proposed framework. In step 1 (fault detection), a convolutional autoencoder (CAE) is trained on the training dataset $\mathcal{D}_{\mathrm{train}}$ containing only healthy data. The discrepancy between the input and reconstructed spectrograms is used to detect faults in the test dataset $\mathcal{D}_{\mathrm{test}}$. In step 2 (fault segmentation), the latent space features in the test dataset $\mathcal{D}_{\mathrm{test}}$ are clustered using clustering algorithms, creating corresponding pseudo-labels. In step 3 (fault diagnostics), an additional classifier $C_{\theta_c}(\cdot)$ is introduced to identify potential fault types based on cluster explanations and is trained using the input latent space features with the generated pseudo-labels as targets. An XAI method is applied to trace the output predictions back to the input spectrograms, providing interpretability and enabling the identification of potential fault types by domain experts.}
\label{fig:framework_ae}
\end{figure*}

The time-series vibration or acoustic signals are first converted into time-frequency spectrograms $\bm{x}$ as inputs to the convolutional autoencoder (CAE). Given a training dataset $\mathcal{D}_\mathrm{train} = \{\bm{x}_0, \bm{x}_1, \dots, \bm{x}_N\}$ consisting of $N+1$ healthy time-frequency spectrograms, the objective is to learn the distribution of healthy data and detect faults in the test dataset $\mathcal{D}_\mathrm{test}$, which may contain both healthy and faulty data. The CAE, a variant of the vanilla autoencoder (AE), is commonly used for tasks such as signal denoising and dimensionality reduction. Unlike the vanilla AE, which uses fully-connected layers, the CAE employs convolutional layers. As illustrated in Figure~\ref{fig:framework_ae}, the CAE consists of two main components: the encoder $E_{\theta_e}(\cdot)$ and the decoder $D_{\theta_d}(\cdot)$. The parameters $\theta_e$ and $\theta_d$ represent the model parameters of the encoder and the decoder, respectively. The encoder compresses the input data into a latent space while retaining essential information, and the decoder reconstructs the original data from these latent features. The CAE is trained on the healthy data to minimize the loss $\mathcal{L}_{\mathrm{CAE}}$, defined as:

\begin{equation}
    \mathcal{L}_{\mathrm{CAE}} =\frac{1}{N+1}\sum_{i=0}^N \left(\bm{x_i} - \hat{\bm{x_i}}\right)^2
\end{equation}

where $\hat{\bm{x_i}}$ is the reconstructed spectrogram from the training sample $\bm{x_i}$, and both $\hat{\bm{x}}_i$ and $\bm{x}_i$ share the same dimensions:

\begin{equation}
    \hat{\bm{x_i}} =  D_{\theta_d}\left(E_{\theta_e}(\bm{x_i})\right).
\end{equation}

\subsection{Fault Detection and Segmentation}
\label{sec:fault_detection_and_segmentation}
% The latent space (embedding) features $\bm{z}$ can be written as 
% \begin{equation}
% \label{eq:latent_space}
%     \bm{z} = E_{\theta_e}(\bm{x})
% \end{equation}

% which will be later used for unsupervised clustering and XAI approach. 
The fault is detected by calculating the residual (reconstruction error) based on the CAE trained solely on healthy data, which means that any deviation from healthy results in a higher residual. Consider an input spectrogram $\bm{x_i} \in \mathbb{R}^{H \times W \times C}$, where $H$ is the spectrogram height, $W$ the width, and $C$ is the number of sensors. The residual $\bm{r_i}$, which has the same shape as the input spectrogram, is defined as:

\begin{equation}
\label{eq:res_ae}
    \bm{r_i} =  \bm{x_i} - D_{\theta_d}(E_{\theta_e}(\bm{x_i})) = \bm{x_i} - \hat{\bm{x_i}}
\end{equation}

where $\hat{\bm{x}}_i$ is the reconstructed spectrogram.

The fault detection identifies faults as samples with residuals exceeding a pre-defined threshold $\tau$. This threshold is calculated based on the residual of the healthy samples in the training dataset $\mathcal{D}_{\mathrm{train}}$, specifically using their mean $\mu$ and standard deviation $\sigma$, as described in Equation~(\ref{eq:threshold}): 

\begin{equation}
\label{eq:threshold}
    \tau = \mu + 3\sigma.
\end{equation}

In our case, the residual $\bm{r}$ has dimensions  $H \times W \times C$. To represent the residual for each sample, we calculate an average over all dimensions, as defined in Equation~(\ref{eq:average_over_image}). A fault in the test dataset $\mathcal{D}_{\mathrm{test}}$ is detected when $\Bar{r} \geq \tau$.
% The mean $\mu$ and standard deviation $\sigma$ are calculated from the average residual $\Bar{r}$.

% over the average value $\Bar{x}$ as defined in Equation~\ref{eq:average_over_image_threshold}. 

% \begin{equation}
% \label{eq:average_over_image_threshold}
%     \Bar{x} = \frac{1}{HWC} \sum_{i=0}^H \sum_{j=0}^W \sum_{k=0}^C \bm{x}_{ijk}
% \end{equation}

% Similar, for a test sample with average residual $\Bar{r}$, it is considered as anomaly if $\Bar{r} > \tau$.

\begin{equation}
\label{eq:average_over_image}
    \Bar{r} = \frac{1}{HWC} \sum_{j=1}^H \sum_{k=1}^W \sum_{l=1}^C \left| r_{jkl}\right|
\end{equation}

where $H$ is the spectrogram height, $W$ the width, and $C$ is the number of sensors. Note that a fixed threshold is used here under the assumption that CBs are operated infrequently, unlike industrial bearings or jet engines, and thus the healthy condition remains stable over time.  Factors such as environmental conditions or interrupted current levels are more likely to influence the distribution of healthy data. For instance, seasonal temperature fluctuations can affect the gas pressure inside CB interruption chambers, thereby impacting contact motion. However, the deviations from faulty to healthy samples are higher than from healthy to healthy under different operation conditions. One way to cope with these deviations within healthy conditions is to include these conditions (such as room temperature, gas pressure, and interrupted current level, etc.) as input features in the framework, allowing the healthy distribution to be dynamically adjusted.

The fault segmentation is achieved through unsupervised clustering methods. It is important to note that the proposed framework is generic and can be used with any clustering method. In this work, we will demonstrate the approach using $K$-means clustering, density-based algorithm OPTICS (Ordering Points To Identify the Clustering Structure)~\cite{ankerst1999optics}, and SOM (Self-Organizing Maps)~\cite{kohonen_som_1990, forest_large-scale_2020}. The number of clusters only has to be specified explicitly for $K$-means but not for the other two methods. The inputs to the clustering methods are the latent space features $\bm{z_i}$ extracted from the trained CAE, as they provide a compressed representation of the input signals. These features, $\bm{z_i}$, can be represented as:
% which aims to minimize the distance between the samples and the centroid of each cluster, thereby determining the centroid of the clusters.
\begin{equation}
\label{eq:latent_space}
    \bm{z_i} = E_{\theta_e}(\bm{x_i}).
\end{equation}

where $E_{\theta_e}(\cdot)$ is the encoder. The dimensionality of these latent space features is a hyperparameter and can vary depending on the specific application.

\subsection{XAI-guided Fault Diagnostics}
\label{sec:xai_fault_diagnostics}
% https://www.tensorflow.org/tutorials/interpretability/integrated_gradients
% As mentioned in Section~\ref{sec:intro}, there are many XAI methods. 
In this step, an attribution-based XAI method is leveraged to generate explanations for each of the resulting clusters. We have adopted the Integrated Gradients (IG) technique in this work because it satisfies two desirable theoretical properties known as completeness and implementation invariance~\cite{forest2024classification}. However, any other XAI technique may be used in our proposed framework.

IG aims to explain model predictions by computing gradients~\cite{sundararajan2017axiomatic}. This method requires a baseline input, typically a black image for image attribution, representing the absence of features contributing to the output. Images are linearly interpolated between the baseline and the input image. Gradients are computed along this path to quantify the relationship between changes in pixel values and changes in the model's prediction. As a result, pixels that contribute more significantly to the model's prediction exhibit higher gradient values.  

Typically, XAI is applied in supervised learning settings, where attribution is calculated from the prediction outputs to input features. However, in our unsupervised learning framework, we only have samples with healthy labels from the training dataset $\mathcal{D}_{\mathrm{train}}$, with no information on fault types. To address this, we create pseudo-labels for the test dataset, which contains both healthy samples and various fault types, based on the clustering results.

To achieve this, we add a $K$-class classifier $C_{\theta_c}(\cdot)$ to the CAE, where $K$ is the number of clusters, as depicted in Figure~\ref{fig:framework_ae}. This network takes as inputs the flattened features from the CAE latent space $\bm{z}$, and classifies them into $K$ classes, as depicted in Equation~(\ref{eq:classifier}). The weights of the encoder $E_{\theta_e}(\cdot)$ are frozen after training the CAE, with only the classifier $C_{\theta_c}(\cdot)$ being trained during this process. The parameters $\theta_c$ represent the classifier parameters.  

% which can be written as 
% \begin{equation}
% \label{eq:latent_space}
%     \bm{z} = E_{\theta_e}(\bm{x})
% \end{equation}

\begin{equation}
\label{eq:classifier}
    \hat{\bm{y}} = C_{\theta_c}(E_{\theta_e}(\bm{x})) =  C_{\theta_c}(\bm{z}) 
\end{equation}

The classifier $C_{\theta_c}(\cdot)$ is trained using samples from the test dataset $\mathcal{D}_{\mathrm{test}}$ as inputs and the one-hot encoded cluster pseudo-labels $\bm{y} \in \{0,1\}^K$ assigned from the clustering results. The training process employs the softmax cross-entropy loss function, as illustrated in Figure~\ref{fig:framework_ae} Step 3. Finally, IG is applied to this classifier to obtain feature attribution explanations for each test sample.

\blue{In this work, the average spectrogram of healthy samples is used as the baseline input in IG. However, due to the sparsity of the attribution maps, interpreting the raw maps remains challenging, even for domain experts. To enhance interpretability, max pooling operations are applied to refine the maps and generate a ``diagnostics matrix". The diagnostics matrix has a lower temporal and frequency resolution, compared to the original attribution maps, making it more accessible for human interpretation due to its lower dimensionality. Each element in the matrix represents the max attribution value for a specific time-frequency region, facilitating a more intuitive fault diagnostics process.} 

\section{Case Study}
\label{sec:experiment}
The International Council on Large Electric Systems (CIGRE)~\cite{Carvalho2012CIGRE} classifies CB malfunctions as “minor” and “major” failures. A CB can still operate when a minor failure occurs, such as some insulating gas leakage. In contrast, the CB operation completely fails due to a major failure. Considering in its investigation CBs from different manufacturers, CIGRE reports that a large proportion of their major failures occur due to malfunctions of the operating mechanism; therefore, the application of monitoring systems with a focus on this module can be very beneficial.

\begin{figure*}[htbp]
\centerline{\includegraphics[width=80mm]{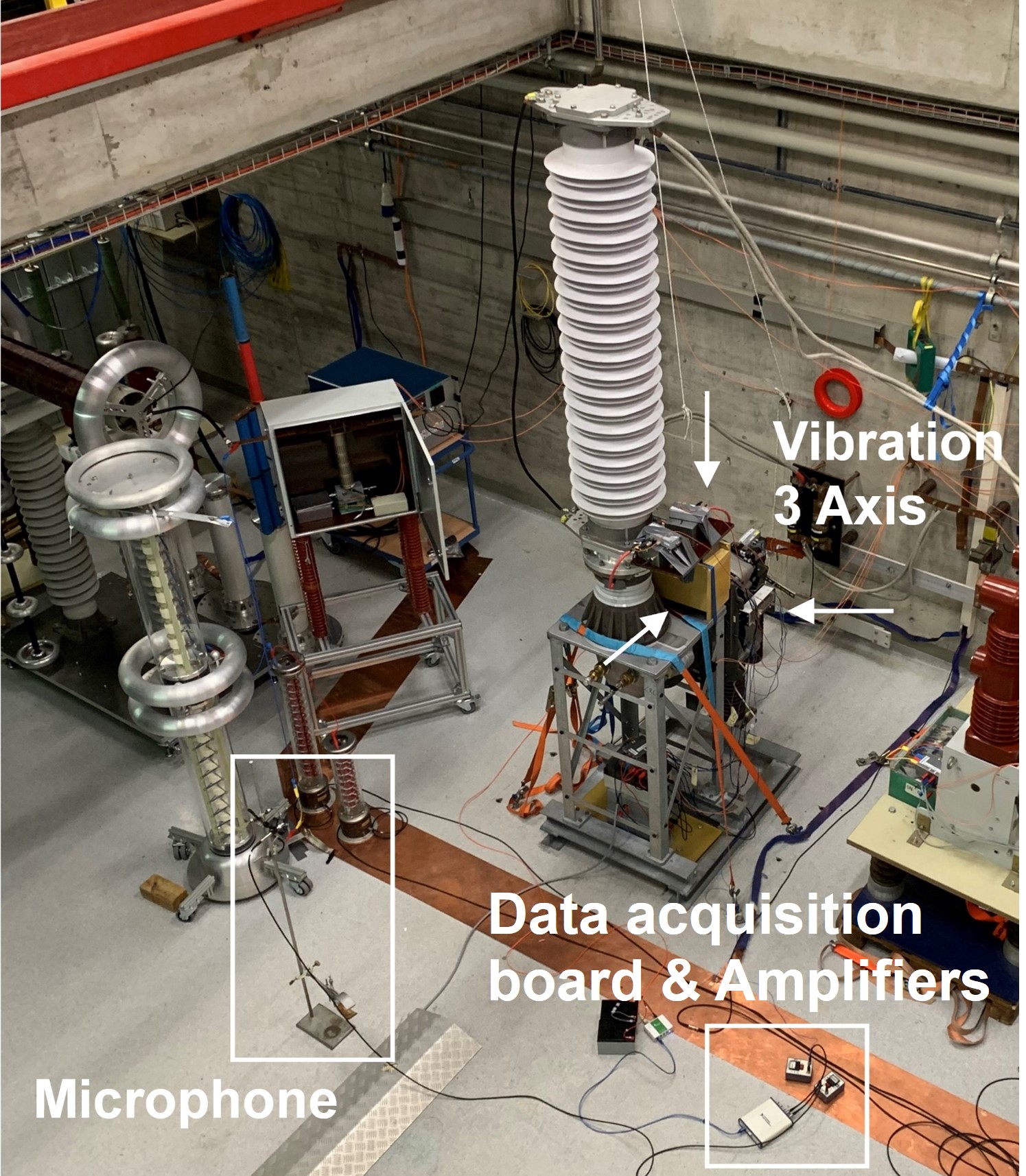}}
\caption{Experimental setup with the test CB, the installation location of microphone, vibration sensors, and the data acquisition board and amplifiers.}
\label{fig:experimental_setup}
\end{figure*}

\begin{figure*}[htbp]
\centerline{\includegraphics[width=130mm]{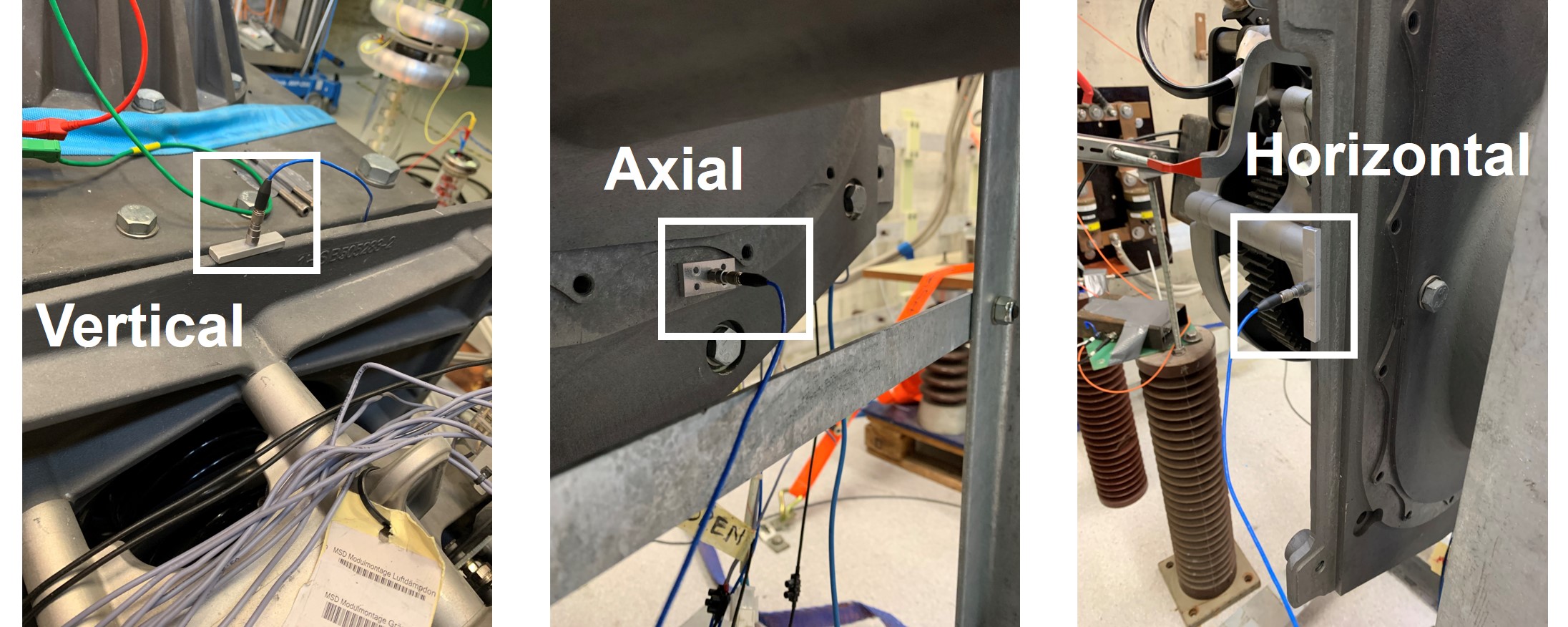}}
\caption{\blue{Installation locations of three vibration sensors (piezoelectric accelerometers) as depicted in Table~\ref{tab:sensors}, adhesive mounted with LOCTITE in three directions with respect to the drive.}}
\label{fig:experimental_setup2}
\end{figure*}

The experimental object was a high-voltage CB operated by a spring drive~\cite{macedo2023diagnostics}, as shown in Figure~\ref{fig:experimental_setup}. The experiment on the CB was conducted under no-load conditions, meaning only mechanical operations without interrupting any current. \blue{To monitor the CB's mechanical dynamics, three piezoelectric accelerometers were installed in three different directions, including horizontal, vertical, and axial, between aluminum mounting bases and the surface of the spring drive structure. The mounting bases have been glued to the drive structure with LOCTITE, and the vibration sensors were tightly screwed to the mounting bases as shown in Figure~\ref{fig:experimental_setup} and Figure~\ref{fig:experimental_setup2}. In addition, a microphone was placed one meter from the drive. Sensor details, including model number, sensitivity, measurement range, and frequency range, are summarized in Table~\ref{tab:sensors}. The installation of these sensors does not affect the integrity or functionality of the test CB as the data are collected non-intrusively.} The data were recorded at a sampling rate of \qty{100}{kS/s} and a sampling length of \qty{2}{s} with two National Instruments boards directly operated with the LabView environment. 
% Several fault types related to springs and dampers were artificially introduced during the experiment. 

\blue{Although the data used to validate the proposed framework were collected in a laboratory setting, we compared it with CB data recorded in a substation and found that noise levels were similar in both environments. This suggests that the performance of the proposed framework should remain robust and unaffected by the typical noisy conditions encountered in substations, provided that sensors and data acquisition systems are properly shielded. Using shielded coaxial or multi-wire cables and ensuring that cable shields are grounded can effectively mitigate interference, maintaining data integrity in real-world deployments.}

\begin{table}[h!]
\centering
\caption{The four sensors used in the experiment and their descriptions. All four sensors are from the same manufacturer PCB Piezotronics. (Acc.: piezoelectric accelerometers)}
\begin{tabular}{|c|c|c|c|c|}
\hline
\textbf{Sensor} & \textbf{Model} & \textbf{Sensitivity} & \textbf{Range} & \textbf{Frequency Range} \\
\hline
Horizontal Acc. & 352A60 & \qty{10}{mV/g} & \qty{\pm 500}{g} & \qtyrange{5}{60000}{\Hz} (\qty{\pm 3}{dB})\\
\hline
Vertical Acc. & M352C18 & \qty{10}{mV/g} & \qty{\pm 500}{g} &  \qtyrange{0.35}{25000}{\Hz} (\qty{\pm 3}{dB})\\
\hline
Axial Acc. & 353B14 & \qty{5}{mV/g} & \qty{\pm 1000}{g} &  \qtyrange{0.35}{30000}{\Hz} (\qty{\pm 3}{dB})\\
\hline
Microphone & 378B02 &  \qty{50}{mV/Pa} & \qty{137}{dB} & \qtyrange{3.75}{20000}{\Hz} (\qty{\pm 2}{dB})\\
\hline
\end{tabular}
\label{tab:sensors}
\end{table}

During the experiment, several fault types related to springs and dampers were artificially introduced. The different combinations of fault types are summarized in Table~\ref{tab:fault_types}. For each condition, blocks of 30 switching operations were conducted. The number of samples was fixed at 30 to be sufficient for reliable statistical analysis while also being a cost-effective compromise. As a result, for each condition, multiples of 30 open and close operations were performed. Some data are missing or additional experiments were performed for Condition \#3, \#4, and \#5, resulting in irregular sample numbers. 

\blue{In real-world scenarios, healthy samples are typically more abundant than faulty ones, which differs from the composition of this dataset. However,  a key advantage of our proposed method is that it does not require any faulty samples during training. The number of faulty samples only influences cluster sizes during the fault segmentation step -- a higher number of faulty samples leads to more distinct clusters. Additionally, cluster imbalance can significantly impact clustering performance. For instance, $K$-means assumes balanced cluster sizes, which may lead to suboptimal results when clusters are highly imbalanced. In contrast, methods such as Gaussian Mixture Models (GMMs) are more flexible, allowing for clusters of varying sizes and densities, making them better suited for imbalanced datasets.}

It is important to note that only 'open' operations were considered in this work, as they are associated with the grid current interruption performance, and therefore, these operations are considered to be more critical. For spring-related faults, under normal conditions, 'normal' spring tension was set as per the high-voltage CB specification, while 'high spring' spring tension was increased to 110\%, and 'low spring' spring tension was reduced to 90\%. Similarly, for damper-related faults, we modified the kinematic viscosity of the damper oil. Under normal conditions, the kinematic viscosity was set at \qty{200}{mm^2/s}. For the faults labeled as 'degraded damper 100' and 'degraded damper 120', the kinematic viscosities were adjusted to \qty{100}{mm^2/s} and \qty{120}{mm^2/s}, respectively.

\begin{table}
    \centering
     \caption{Spring and damper conditions in the dataset. Only Condition \#1 is considered as healthy, while all others are faulty. Two sub-conditions from Condition \#2 (nSdD100 and nSdD120) are combined into nSdD and other two from condition \#5  (lSdD100 and lSdD120) are lSdD.}
    \begin{tabular}{|c|c|c|c|c|c|}
    \hline
       Condition \# & Spring & Damper & Notation & \# samples\\
    \hline
       1  & normal & normal & nSnD & 60 \\
    \hline
       \multirow{2}{*}{2}  & \multirow{2}{*}{normal} & degraded 100 & nSdD100 & 30 \\
         &  & degraded 120 & nSdD120 & 30 \\
    \hline
       3  & high & normal & hSnD & 96 \\
    \hline
       4  & low & normal & lSnD & 65 \\
    \hline
       \multirow{2}{*}{5}  & \multirow{2}{*}{low} & degraded 100 & lSdD100 & 29 \\
         & & degraded 120 & lSdD120 & 30 \\
    \hline
    \end{tabular}
   
    \label{tab:fault_types}
\end{table}

% The ground-truth labels are defined as follows: 0: low spring normal damper, 1: high spring normal damper, 2: normal spring normal damper, 3: normal spring degraded damper, 4: low spring degraded damper.

\begin{figure*}[htbp]
\centerline{\includegraphics[width=\textwidth]{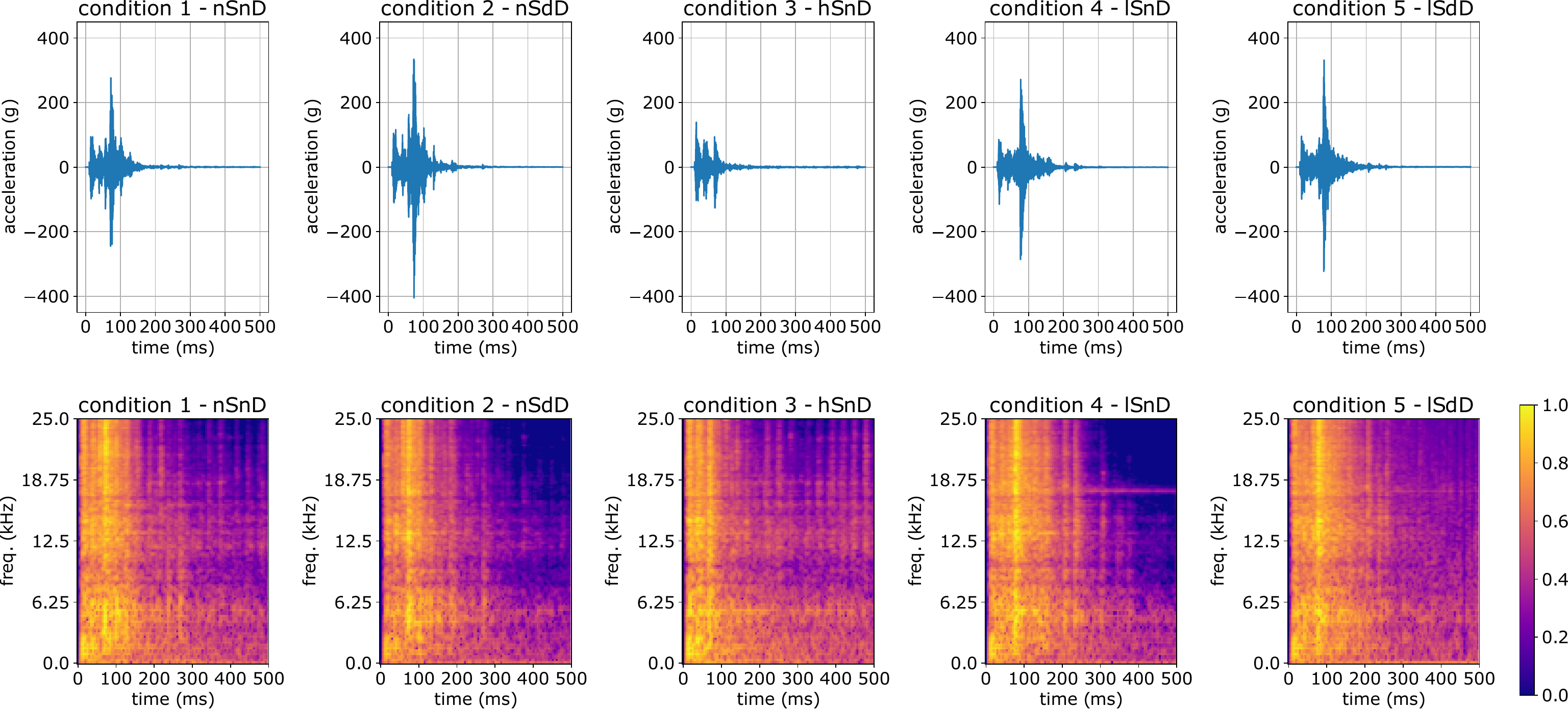}}
\caption{Example of vibration signals in the vertical direction for all five conditions (upper row) and their corresponding Mel spectrograms (bottom row). The samples shown here for Condition \#2 is nSdD120 and for Condition \#5 is lSdD120.}
\label{fig:example_vibration}
\end{figure*}

\subsection{Data Pre-processing}
% \subsection{Short-time Fourier Transform (STFT)}
The signals from the four sensors are recorded simultaneously. Manual inspection revealed that most of the vibration signals are concentrated in the first \qty{500}{ms}, with vibrations damping out thereafter. Therefore, to reduce computational costs, only the first \qty{500}{ms} were retained for analysis. For input into the CAE, log-Mel spectrograms, a type of time-frequency spectrogram, were extracted from the truncated signals, as Mel spectrograms have been proven effective in many applications such as detecting malfunctions in industrial machinery~\cite{purohit2019mimii} and recognizing speech emotions~\cite{meng2019speech}. Time-frequency spectrograms were used because they facilitate the extraction and analysis of information. They provide a two-dimensional representation of a signal, displaying both frequency and time information simultaneously. By converting a time-series signal into its time-frequency components, spectrograms make it easier to detect and analyze features that are not readily apparent in the time-domain alone.

Given the presence of multiple sensors, spectrograms from each sensor were concatenated channel-wise. This concatenation ensures that each pixel in the spectrogram contains comprehensive information at a specific time and frequency across all sensors, assuming synchronized data acquisition. As mentioned in Section~\ref{sec:fault_detection_and_segmentation}, the input spectrogram has a size of $H \times W \times C$. In this case, $H$ represents the number of frequency bins, $W$ is the number of time bins, and $C$ is the number of sensors. Specifically, the input dimensions were set to  $128 \times 100 \times 4$ in this work. Example vibration signals in the vertical direction for Condition \#1 to \#5 and their corresponding Mel-spectrograms are shown in Figure~\ref{fig:example_vibration}. Note that the sample for Condition \#3 shows low vibration amplitudes compared to samples with different conditions and is thus simple to recognize, whereas the other four conditions are challenging to distinguish visually. 

% , where the vibration signals are normalised based on the highest absolute amplitude within the five signals.

% we have is the Mel spectrogram time-frequency maps, where the vibration signals are divided into different segments and Fourier transform is performed on each segment. In addition, we have more than one vibration sensors, the time-frequency maps extracted from the additional sensors are concatenated channel-wise. In the end, the input data has a size of $T \times F \times C$, where $T$ is the number of time bins, $F$ number of frequency bins, and $C$ number of channel (number of sensors). 

\subsection{Evaluation Metrics}
\label{sec:evaluation}
% https://scikit-learn.org/stable/modules/clustering.html#clustering-evaluation
Four metrics are used to evaluate the clustering performance with respect to the ground-truth labels. The first one is the adjusted Rand index (ARI)~\cite{hubert1985comparing}. The ARI measures the similarity between two clustering results, offering a normalized score that adjusts for chance agreement. Specifically, it compares the clustering generated by the algorithm, denoted as $\mathcal{C}$, with a ground-truth clustering $\mathcal{K}$, where correct class assignments are known. The underlying Rand Index (RI) quantifies the agreement between these two clusterings by considering all pairs of elements and counting pairs that are either assigned to the same or different clusters in both $\mathcal{K}$ and $\mathcal{C}$. The ARI adjusts this measure to account for the chance grouping of elements, thus providing a more accurate assessment of the clustering validity. The RI is defined as follows:
%After we have the fault segmentation clustering results from the k-means algorithm, we compare it with the ground-truth fault types, which has three different fault types and each fault type contains 10 units. The evaluation metric used here is the adjusted Rand Index (ARI), which calculates the similarity between two clustering results by comparing pair-wise samples. If C is the ground truth cluster with a correct class assignment and K is the cluster we want to investigate, the unadjusted Rand Index (RI) is calculated as 
\begin{equation}
   \text{RI} = \frac{a+b}{C^{n_\text{samples}}_2}
\end{equation}
where $a$ is the number of pairs of samples that are placed in the same cluster in both $\mathcal{K}$ and $\mathcal{C}$, and $b$ is the number of pairs that are placed  in different clusters in both clusterings. The denominator, $C^{n_\text{samples}}_2$ is the total number of possible pairs in the dataset and $n_\text{samples}$ is the number of samples. The ARI is the modified version of RI to correct for the  chance grouping of elements. It is defined as follows: 
\begin{equation}
   \text{ARI} = \frac{\text{RI}-\mathbb{E}(\text{RI})}{\text{max(RI)}-\mathbb{E}(\text{RI})}
\end{equation}

where $\mathbb{E}(\text{RI})$ is the expected value of the RI under random classification. An ARI of 1  indicates perfect agreement between the  two clustering results relative to chance,  while an ARI of 0 suggests that the clustering is no better than random.

The other three metrics are the homogeneity score $h$, completeness score $c$, and v-measure $v$~\cite{rosenberg2007v}, which are commonly used in unsupervised clustering. The homogeneity score $h$ (also known as purity) measures how well each cluster contains only data points from a single class. It is defined as: 
\begin{equation}
\label{eq:homogeneity}
    h = 1 - \frac{H(\mathcal{C}|\mathcal{K})}{H(\mathcal{C})}
\end{equation}
where $H(\mathcal{C})$ is the entropy of the data classes and $H(\mathcal{C}|\mathcal{K})$ is the conditional entropy of the classes  given the cluster assignments. Similarly, the completeness score $c$ measures how well all data points of a particular class are assigned to the same cluster. It is defined as: 
\begin{equation}
    c = 1 - \frac{H(\mathcal{K}|\mathcal{C})}{H(\mathcal{K})}
\end{equation}
where $H(\mathcal{K})$ is the entropy of the cluster assignments and $H(\mathcal{K}|\mathcal{C})$ is the conditional entropy of the cluster assignments given the data. The v-measure $v$ combines both homogeneity and completeness and is defined as: 
\begin{equation}
    v=\frac{2\cdot h \cdot c}{h+c}.
\end{equation}
All three metrics range between 0 and 1, with a score of 1 indicating perfect clustering results.

To quantitatively evaluate the SOM results, we also report internal metrics. Three metrics are used to validate that the SOM accurately represents the data distribution and has good topological organization~\cite{forest_survey_2020}. These metrics include: quantization error, topographic error, and topographic product. Quantization error measures the average Euclidean distance error introduced when projecting data onto the SOM, while topographic error evaluates the neighborhood preservation of the projection (lower is better). The topographic product assesses the smoothness and preservation of neighborhood relations between the SOM map and the input space, where values closer to 0 are better.
% well as the class scatter index, which assesses whether samples of the same ground-truth class are grouped in neighboring regions of the map.

For the performance of the XAI results, we utilize the concept of faithfulness~\cite{alvarez2018towards}, which measures how accurate are the features highlighted by the attribution explanations. To do so, faithfulness evaluation involves measuring the change in the classifier $C_{\theta_c}(\cdot)$ output when occluding the features selected by an explanation (typically replacing them with zeros). In our case, the features correspond to individual pixels in the spectrograms. First, in order to assess the quality of the explanations obtained with an attribution method (such as Integrated Gradients), we perform an attribution-based occlusion, where the features with the highest attribution values (the most important ones according to the XAI method) are occluded. Then, we perform a random occlusion, where sets of features are randomly selected and occluded. If the explanations are meaningful, attribution-based occlusion should lead to larger changes in the model output than random occlusion. As the spectrograms have a high dimension, features are not replaced one by one, but by groups for each channel. The modified spectrograms $\bm{\tilde{x}}$ are then fed into the trained encoder and classifier. For each occlusion, we compute the absolute difference in the classifier's outputs for the predicted class before and after the occlusion, as defined in Equation~\ref{eq:faithfulness}, for each sample. The outputs are taken from the last layer after applying the softmax activation function, providing insights into how the removed features impact the model's confidence in the prediction:

\begin{equation}
\label{eq:faithfulness}
    \Delta \text{prediction}(\bm{x}) = \left| C_{\theta_c}(E_{\theta_e}(\bm{x}))[\hat{y}] -  C_{\theta_c}(E_{\theta_e}(\bm{\tilde{x}}))[\hat{y}] \right|,
\end{equation}

where $\hat{y} = \arg\max C_{\theta_c}(E_{\theta_e}(\bm{x}))$ is the predicted class for the original input.

\subsection{Model Architecture and Hyperparameter Settings}
To calculate the Mel spectrogram, the number of Mel bands was set to 128, and the hop length was set to 501, resulting in a spectrogram for one operation with four sensors of size $128 \times 100 \times 4$. The CAE architecture is summarized as follows: the encoder $E_{\theta_e}(\cdot)$ consists of Conv2D ($16 \times 3 \times 3$), Max Pooling ($2 \times 2$), Conv2D ($8 \times 3 \times 3$), Max Pooling ($2 \times 2$), Conv2D ($1 \times 3 \times 3$), Max Pooling ($2 \times 5$). Here, Conv2D ($f \times f_x \times f_y$) represents a 2D convolutional layer with $f$ filters and a filter size of $f_x \times f_y$, and Max Pooling ($m_x \times m_y$) represents the max pooling layer with a size of $m_x \times m_y$. Similarly, the decoder $D_{\theta_d}(\cdot)$ has the reversed architecture as the encoder $E_{\theta_e}(\cdot)$, but instead of 2D convolutional layers, it uses deconvolutional layers, and instead of max pooling layers, it uses 2D up-sampling layers. This architecture is selected from a grid search. With this architecture and the shape of input Mel spectrogram $128 \times 100$, the latent space has a dimension of $16 \times 5$. The activation functions in the CAE are all rectified linear units (ReLU), except for the final output layer, which uses a linear activation.

The classifier $C_{\theta_c}(\cdot)$ has a simple architecture comprising only one layer. The input layer is the flatten layer of the latent space with 80 neurons, which is the size of the flattened CAE latent space. The output is a fully connected layer with 5 neurons, representing one-hot encoded $K=5$ clusters identified from $K$-means. The activation function is softmax, and no bias is applied in the classifier. \blue{To generate a diagnostics matrix, max pooling operations with a pooling size of (32, 20) are applied to the attribution maps, which have a same shape as the spectrogram. The diagnostics matrix, in this case, has a temporal resolution of five intervals (interval of \qty{100}{ms}) and a frequency resolution of four bands (low, mid-low, mid-high, high).}

The CAE was trained for 300 epochs with a batch size of 8, using early stopping with patience of 10 epochs. The epoch indicates how many times the training data is fed through the model. The Adam optimizer~\cite{kingma2014adam} with $\beta_1 = 0.9$ and $\beta_2 = 0.999$, a learning rate of 0.001, and the mean squared error loss function were used. The CAE was trained only on the healthy data (Condition \#1 - nSnD) defined in Table~\ref{tab:fault_types}. A randomly selected 10\% subset of the healthy data served as the validation dataset. The classifier $C_{\theta_c}(\cdot)$ was trained on the test dataset using the same training procedure as the CAE after fault segmentation step, but with categorical cross-entropy as the loss function.

For the unsupervised clustering analysis, the number of clusters $K$ for $K$-means is set to 5 using elbow curve analysis, initialized with $K$-means++. For the OPTICS algorithm, the neighborhood size is set to 5 samples, and the distance parameter $p$ is set to 2, corresponding to the Euclidean distance. Both $K$-means and OPTICS are implemented using scikit-learn. For the SOM, a $10 \times 10$ grid is used, with a Gaussian neighborhood function, a spread $\sigma$ of 5, and a learning rate of 0.05. The SOM implementation is using the MiniSom~\cite{vettigliminisom}, and performance metrics using SOMperf~\cite{forest_survey_2020}. All clustering algorithms operated on the 80-dimensional latent space features extracted from the trained CAE.

%% file: results.tex
\section{Results and Discussions}
\label{sec:results}
In this section, the performance of the proposed fault detection and segmentation framework is analyzed. First, faulty samples are identified from the collected CB data through the fault detection process. Next, these samples are grouped into different clusters for fault segmentation without requiring prior knowledge of the specific fault types. Finally, an XAI approach is applied to interpret the clusters, providing insights for XAI-guided fault diagnostics.  

While the results presented here are based on a single CB, we assume that the domain gap between different CBs of the same type is relatively small compared to the differences between healthy and faulty data. This assumption is based on the fact that high-voltage CBs in gas-insulated switchgear are common to have one CB per phase. In three-phase systems, the three CBs or even all CBs in a substation are generally of the same type, installed and commissioned at the same time, and positioned next to each other, leading to similar operating conditions and histories. The data collected from these CBs could be combined and used as inputs to the proposed framework.

\subsection{Fault Detection - CAE}
\label{sec:results_anomaly}
\begin{figure}[htbp]
\centerline{\includegraphics[width=100mm]{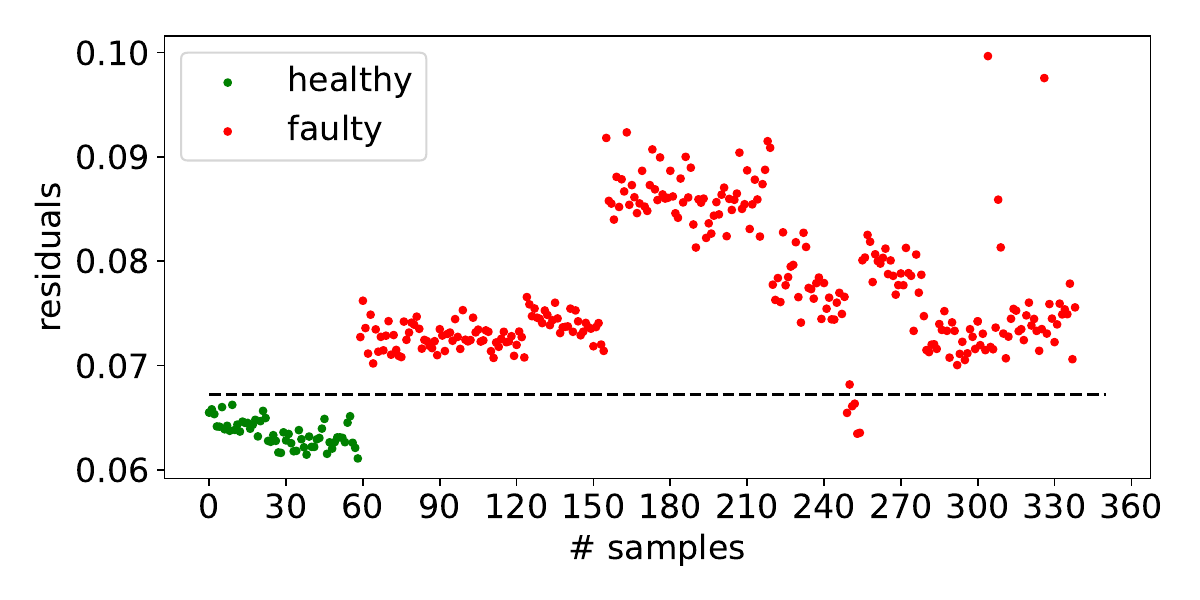}}
\caption{The fault detection results based on the CAE residuals with healthy (green) and faulty samples (red) from the ground truth. Horizontal dashed line represents the threshold $\tau$ defined in Equation~(\ref{eq:threshold}).}
\label{fig:anomaly_detection}
\end{figure}

The CAE residuals as defined in Equation~(\ref{eq:average_over_image}) are plotted in Figure~\ref{fig:anomaly_detection}, where healthy samples are colored in green, faulty in red, based on the ground-truth. A horizontal dashed line represents the threshold $\tau$ defined in Equation~(\ref{eq:threshold}) based on the healthy data. False negative samples occur when faulty samples (red) have residuals below the threshold, indicating that the model fails to detect these faults. In this case, the false negative rate is 1.79\%.  Overall, approximately 98.21\% of faulty samples can be detected with the CAE trained only on the healthy samples.

% Additionally, We observe different in residual levels from Figure~\ref{fig:anomaly_detection}. For instance, residuals slightly exceed the threshold from approximately sample \# 60 to \# 160 and from sample \# 280 to \# 350, while residuals are higher from sample \# 160 to \# 220, suggesting potential differences in fault types. To separate the detected anomalies, we applied unsupervised K-means clustering and the results are shown in Section~\ref{sec:results_clustering}.  

\begin{figure}[htbp]
\centerline{\includegraphics[width=140mm]{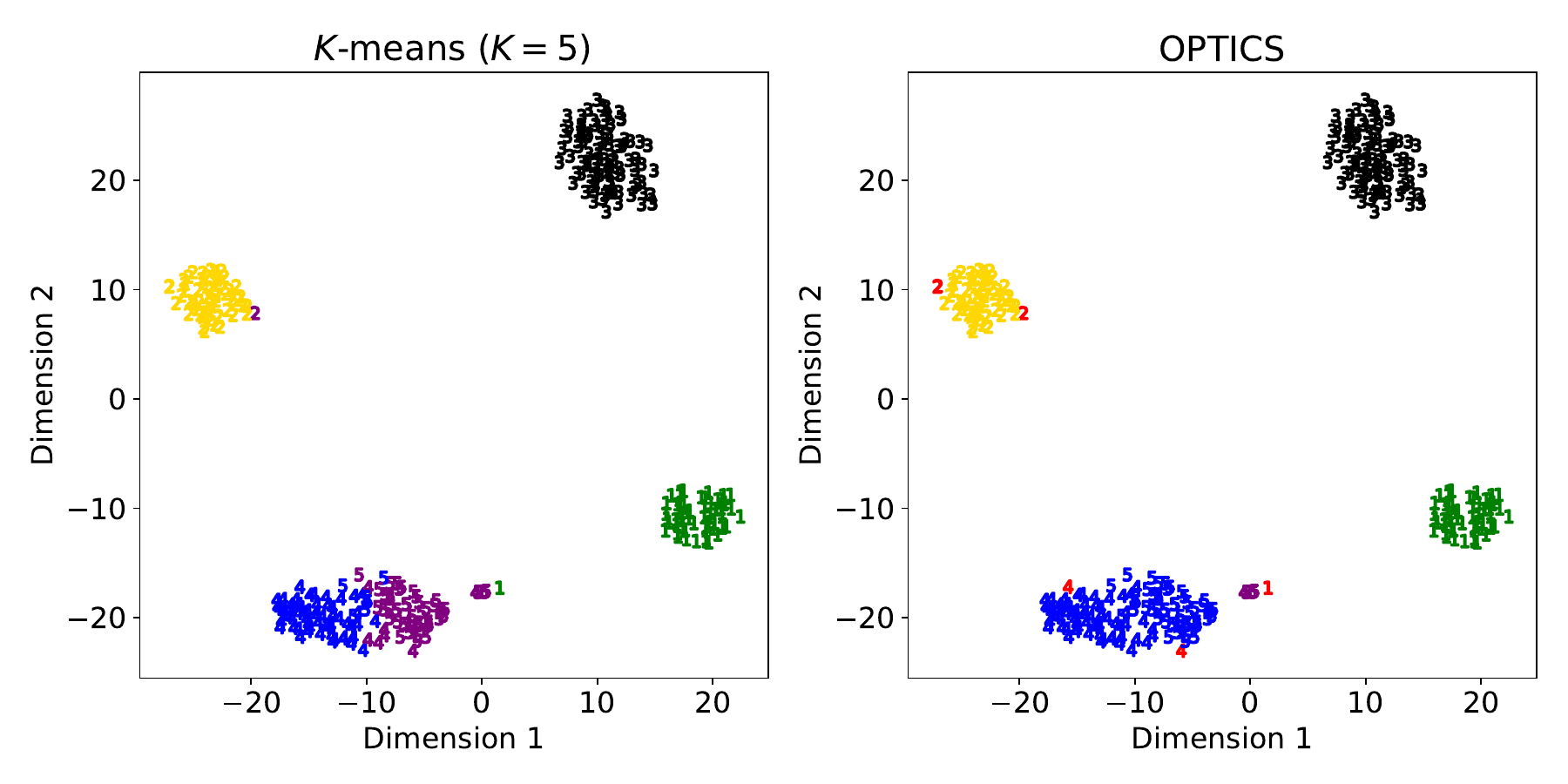}}
\caption{Offline clustering results using $K$-means (left) and OPTICS (right) with the CAE latent features, visualized in 2D using $t$-SNE. Colors are assigned from the clustering results, while markers are the ground-truth conditions as defined in Table~\ref{tab:fault_types}: 1 - nSnD, 2 - nSdD, 3 - hSnD, 4 - lSnD, 5 - lSdD. Red samples in OPTICS clustering results correspond to the outliers.}
\label{fig:results_clustering_tsne}
\end{figure}

% \begin{figure}[htbp]
% \centerline{\includegraphics[trim={1cm 2.6cm 1cm 1cm},width=100mm]{clustering_k5_4064_mt0_100_mf0_128_n_iter1.pdf}}
% \caption{Unsupervised clustering results using $K$-means with the CAE latent space features and $K=5$. Each axis represents a PC. The colors are assigned from the $K$-means results, while the markers are the ground-truth conditions as defined in Table~\ref{tab:fault_types}: 1 - nSnD, 2 - nSdD, 3 - hSnD, 4 - lSnD, 5 - lSdD.}
% \label{fig:results_clustering}
% \end{figure}
% The ground-truth labels are defined as follows: 0: low spring normal damper, 1: high spring normal damper, 2: normal spring normal damper, 3: normal spring degraded damper, 4: low spring degraded damper. The number of clusters $k$ is selected to be five based on the elbow curve. 

\subsection{Fault Segmentation - Offline Clustering}
For simplicity, we first consider an offline clustering setting where the full dataset is available. Clustering results obtained by $K$-means and OPTICS are shown in Figure~\ref{fig:results_clustering_tsne}, using $t$-Distributed Stochastic Neighbor Embedding ($t$-SNE) to map the high-dimensional data into a two-dimensional space. Colors represent the clusters identified by the clustering algorithm, while marker numbers correspond to the ground-truth labels defined in Table~\ref{tab:fault_types}. For $K$-means, the black, green, and yellow clusters exhibit clear separation, while the blue and purple clusters show slight overlap. Some samples are incorrectly assigned to other clusters (e.g., purple 4s or blue 5s). However, both blue and purple clusters primarily contain samples in Condition \#4 and \#5, representing low spring faults and differing only in damper conditions. OPTICS (on the right of Figure~\ref{fig:results_clustering_tsne}), however, failed to distinguish between these two clusters and assigned all samples from these two conditions to a single large cluster and a small cluster. No additional sub-clusters are visible within the blue and purple clusters, indicating that it is challenging to further separate the fault sub-types such as different levels of degraded damper (between kinematic viscosity \qty{100}{mm^2/s} and \qty{120}{mm^2/s}) as described in Table~\ref{tab:fault_types}.

%  and demonstrate the flexibility of the proposed framework on clustering algorithms
% When compared to the ground-truth labels, the samples with labels 1, 2, and 3 exhibit clear separation compared to the samples with labels 4 and 5. Some samples from classes 4 and 5 are incorrectly assigned to the other clusters (e.g., purple 4s or blue 5s). 

% In addition, we observe a segregation between high (black), normal (green and yellow), and low (blue and purple) spring conditions along the PC1 direction. This implies that the spring conditions could have a greater influence on the CB vibration and acoustic signals than the damper conditions. Conversely, the model struggles to distinguish between different damper conditions.

Clustering performance metrics, including the ARI score, homogeneity score $h$, completeness score $c$, and v-measure $v$, as described in Section~\ref{sec:evaluation}, are summarized for both clustering methods in Table~\ref{tab:clustering_metrics}. $K$-means has higher scores across all four metrics, with all four metrics exceeding $0.9$. However, OPTICS has lower scores because of the misclassification of two different damper conditions under low spring conditions. Despite this, OPTICS offers an advantage over $K$-means: it does not require specifying the number of clusters, which is beneficial in real-world applications, where the number of clusters is typically unknown.

% In real-world applications, the selection of the optimal clustering methods depends on the applications. In scenarios involving the monitoring of only a few CBs, the maximum number of clusters could be estimated as the same as the number of CBs (assume one fault type per CB). In practice, the actual number is likely to be lower, as the probability of all CBs experiencing unique fault types simultaneously is rare. In this case, $K$-means is preferred as the number of clusters $K$ is low and easy to infer. Conversely, in the case of monitoring 100 or even more CBs at the same time, determining the number of clusters becomes challenging. In this case, clustering methods such as OPTICS or SOM, which do not require a predefined number of clusters, are better suited despite the potential trade-off in clustering performance.

\begin{table}[h]
    \centering
        \caption{Clustering performance of the $K$-means and OPTICS clustering methods. The symbol $\uparrow$ means the higher the value is, the better separated the clusters. The best score is 1 for all four metrics, where clusters are well separated.}
    \begin{tabular}{|c|c|c|c|c|}
    
    \hline
        Clustering method & ARI $\uparrow$ & $h$ $\uparrow$ & $c$ $\uparrow$ & $v$ $\uparrow$\\
    \hline
         $K$-means & \textbf{0.9172} & \textbf{0.9137} & \textbf{0.9136} & \textbf{0.9137}\\
    \hline
        OPTICS & 0.8018 & 0.8366 & 0.8996 & 0.8670\\
    % \hline
    %     SOM & 0. & 0. & 0. & 0.\\
    \hline
    \end{tabular}

    \label{tab:clustering_metrics}
\end{table}

The confusion matrix of the $K$-means results is shown in Table~\ref{tab:confusion_matrix} as it achieves the highest scores in Table~\ref{tab:clustering_metrics}. As shown in Figure~\ref{fig:results_clustering_tsne}, the majority of misclassified samples occur between Condition \#4 (lSnD) and \#5 (lSdD). It is important to note that the clustering results do not inherently indicate which cluster corresponds to which fault label. This confusion matrix serves only for evaluation purposes, as in real-world scenarios ground-truth labels are unavailable.
% based on the group truth, which is unavailable in real-world scenarios. 
% , but it is challenging to distinguish between different level of degraded damper conditions.

\begin{table}[h]
    \caption{Confusion matrix of the $K$-means clustering results with $K=5$. Note that the ground-truth labels are unavailable in real-world scenarios, and one does not know which cluster corresponds to which condition.}
    \centering
    \begin{tabular}{|c|l|c|c|c|c|c|}
    \hline
        \multicolumn{2}{|c|}{} & \multicolumn{5}{c|}{Prediction} \\
    \cline{3-7}
        \multicolumn{2}{|c|}{} & 1 - nSnD & 2 - nSdD & 3 - hSnD & 4 - lSnD & 5 - lSdD \\
    \hline
        \multirow{5}{*}{Actual}& 1 - nSnD & 60 & 0 & 0 & 0 & 0\\
    \cline{2-7}
        & 2 - nSdD & 0 & 59 & 0 & 0 & 1\\
    \cline{2-7}
        & 3 - hSnD & 0 & 0 & 96 & 0 & 0\\
    \cline{2-7}
        & 4 - lSnD & 0 & 0 & 0 & 58 & 7\\
    \cline{2-7}
        & 5 - lSdD & 0 & 0 & 0 & 6 & 53\\
    \hline
    
    \end{tabular}

    \label{tab:confusion_matrix}
\end{table}
% degrees of degraded damper conditions 

\begin{figure}[htbp]
\centerline{\includegraphics[width=100mm]{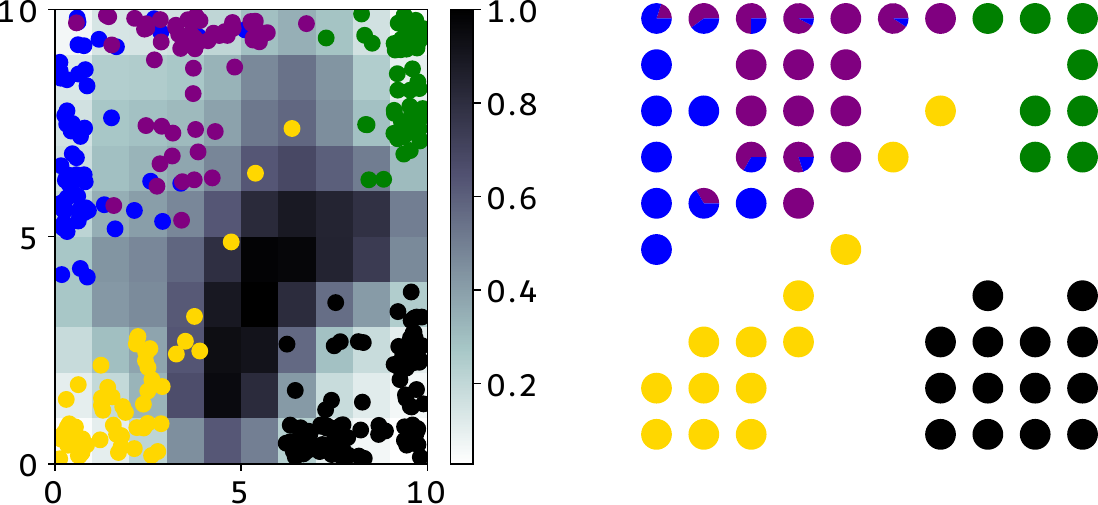}}
\caption{Offline clustering results using SOM with grid size (10, 10), colored by the ground-truth labels. On the left, the U-matrix indicates the distance between neighboring cells with the projected samples. On the right, the SOM cells are colored by class assignments.}
\label{fig:clustering_som_offline}
\end{figure}

Besides $K$-means and OPTICS, the clustering performance of the SOM with grid size $(10, 10)$ is shown in Figure~\ref{fig:clustering_som_offline} for the entire dataset in an offline setting. On the left of Figure~\ref{fig:clustering_som_offline}, the U-matrix is presented, indicating the distance between neighboring cells with the projected samples. On the right side, the SOM cells are colored by class assignments. Quantitative evaluation of the SOM was performed using metrics discussed in Section~\ref{sec:evaluation}. The homogeneity score $h$ is $0.9564$, higher than $K$-means and OPTICS as shown in Table~\ref{tab:clustering_metrics}. The map properly approximates the data distribution and is well-organized, with quantization and topographic errors of $0.1383$ and $0.0559$ respectively, and a topographic product equal to $0.0101$, close to zero.

% \begin{table}[h]
%     \centering
%         \caption{\blue{SOM clustering performance. The symbol $\uparrow$ means the higher the value is, the better separated the clusters.}}
%     \begin{tabular}{|c|c|c|c|c|}
    
%     \hline
%          & $h$ $\uparrow$ & quantization errors $\downarrow$ & topological errors $\downarrow$ & topological products $0$\\
%     \hline
%          SOM & 0.9564 & 0.1383 & 0.0559 & 0.0101\\
%     \hline
%     \end{tabular}

%     \label{tab:som_clustering_metrics}
% \end{table}

% \begin{figure}[htbp]
% \centerline{\includegraphics[width=70mm]{figures/xai_query_sample_4064_mt0_100_mf0_128_n_iter1_channel2.pdf}}
% \caption{The representative instances for each condition, which has the shortest distance in PC space to the cluster centroid. The vibration in axial direction (upper row) and spectrograms (middle row) signals and the corresponding IGs (lower row) are shown for each condition. The signals are normalised to the highest peak within all conditions.}
% \label{fig:query_samples_axi}
% \end{figure}

\subsection{Fault Segmentation - Online Clustering}
\label{sec:results_clustering}

% To illustrate the clustering results, principal component analysis (PCA) is applied to the CAE latent space features of dimension 80. The first three principal components (PCs) are used for visualization in a 3D plot, with each axis representing a PC, as shown in Figure~\ref{fig:results_clustering_tsne}.

\begin{figure*}[htbp]
\centerline{\includegraphics[width=\textwidth]{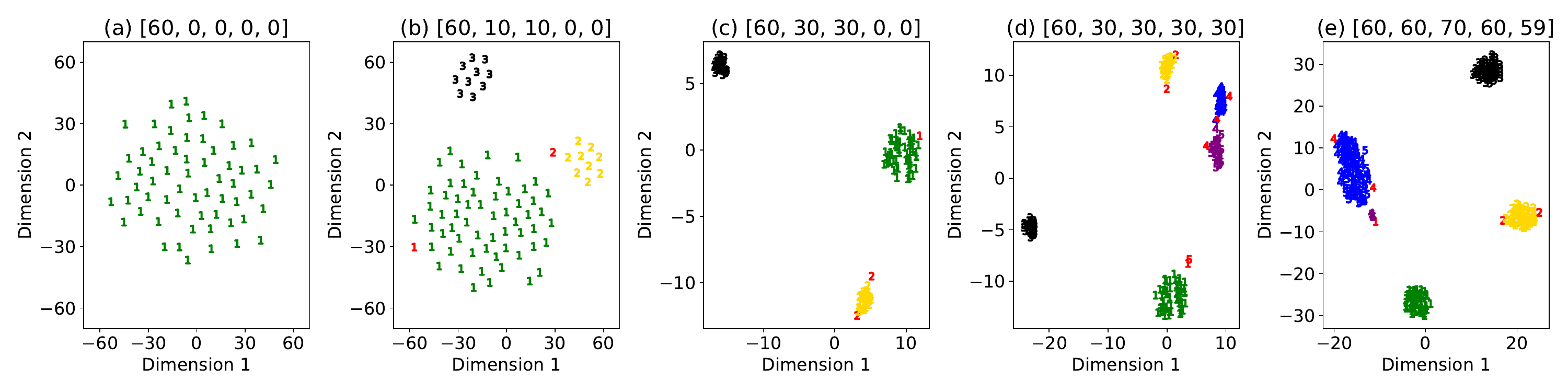}}
\caption{Clustering results using OPTICS with the CAE latent features over an incremental data stream (from (a) to (e)), visualized in 2D space using $t$-SNE. Colors are assigned from the clustering results, while markers are the ground-truth conditions as defined in Table~\ref{tab:fault_types}: 1 - nSnD, 2 - nSdD, 3 - hSnD, 4 - lSnD, 5 - lSdD. The subfigure titles indicate the number of samples available in each class at each step of the data stream. For example, in (a) we have 60 samples from Condition\#1 (nSnD).}
\label{fig:results_clustering_over_time}
\end{figure*}

\begin{figure*}[htbp]
\centerline{\includegraphics[width=\textwidth]{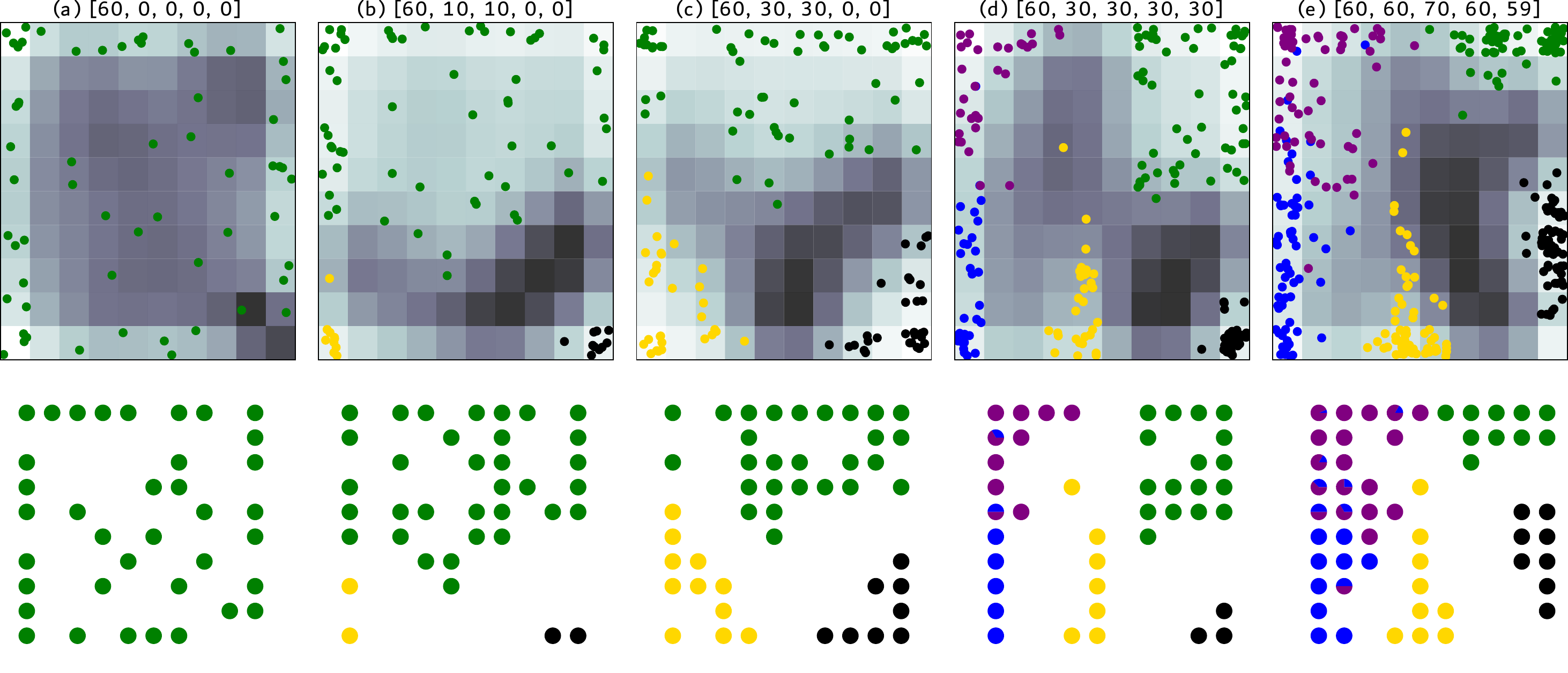}}
\caption{Clustering results using SOM with the CAE latent features over an incremental data stream (from (a) to (e)). Top row: U-matrix (indicating distance between neighboring cells) with projected data samples. Bottom row: map cells colored by class assignments. Colors are assigned from the ground-truth conditions as defined in Table~\ref{tab:fault_types}: green - nSnD, yellow - nSdD, black - hSnD, blue - lSnD, purple - lSdD. The subfigure titles indicate the number of samples available in each class at each step of the data stream.}
\label{fig:clustering_som_over_time}
\end{figure*}

In real-world applications, the complete dataset is not available initially in most cases, and new data are rather being incrementally streamed from substations with each CB switching operation. Thus, we consider an online clustering setting where new fault types potentially appear over time. In this setting, OPTICS and SOM are more suitable as they can better adapt to new incoming data, unlike $K$-means, which requires the number of clusters to be predefined at each step. In Figure~\ref{fig:results_clustering_over_time} and Figure~\ref{fig:clustering_som_over_time}, we depict the online clustering results for OPTICS and SOM over time. The title of each subfigure denotes the data distribution available for each label defined in Table~\ref{tab:fault_types}. The samples available at each time step of the data stream are assigned arbitrarily.
%In Figure~\ref{fig:clustering_som_over_time}, the upper rows correspond to the }
% It is important to emphasize that these ground-truth labels are unknown to the clustering model and are also unavailable in real-world applications.} 

% check tense for this paragraph
Initially, the CAE is trained with only healthy data, resulting in the formation of a single cluster, as depicted in Figure~\ref{fig:results_clustering_over_time} (a) and Figure~\ref{fig:clustering_som_over_time} (a). Subsequently, 20 new samples were collected, consisting of 10 Condition \#2 (nSdD) and 10 Condition \#3 (hSnD) samples. As shown in Figure~\ref{fig:results_clustering_over_time} (b) and Figure~\ref{fig:clustering_som_over_time} (b), both clustering algorithms identified two new clusters (black and yellow), potentially indicating two distinct fault types. Notably, some samples are marked in red in the case of OPTICS, representing outliers that the clustering algorithm could not assign to any cluster. Later, another batch of 40 samples (20 additional samples for Condition \#2 and \#3 each) was collected and the clustering results are shown in Figure~\ref{fig:results_clustering_over_time} (c) and Figure~\ref{fig:clustering_som_over_time} (c).

Interestingly, in Figure~\ref{fig:results_clustering_over_time} (d), the cluster for Condition \#1 splits into two sub-clusters, even though they are still considered as a single cluster by the clustering algorithm. After inspecting the ground-truth labels, it becomes evident that these sub-clusters correspond to two experimental blocks conducted on different days. This suggests that the healthy condition can deviate because of varying experimental conditions such as room temperature and gas pressure. However, the deviations between faulty and healthy conditions are considerably larger compared to the variations observed among healthy samples.

As shown in Figure~\ref{fig:results_clustering_over_time} (d) and Figure~\ref{fig:clustering_som_over_time} (d), new faulty samples result in two new clusters. However, in Figure~\ref{fig:results_clustering_over_time} (e) and the clustering results of the full dataset on the right of Figure~\ref{fig:results_clustering_tsne}, the purple and blue clusters merge into a single large blue cluster and a small purple cluster. Similarly, the purple and blue clusters are close to each other and do not have a clear boundary in Figure~\ref{fig:clustering_som_over_time} (e). \blue{This demonstrates that even with a limited number of faulty samples from the online data streaming process -- a scenario commonly encountered in real-world applications -- the proposed framework remains effective. It can still successfully segment most faulty samples into their corresponding clusters, ensuring reliable fault identification even under data scarcity.}

\blue{We have demonstrated the feasibility of using different clustering methods with our proposed framework. The selection of the optimal clustering methods in real-world applications depends on the specific use case as different algorithms highlight different cluster properties. Cluster analysis is usually an interactive process, where users explore the underlying structure of the data distribution. Due to its unsupervised nature, defining a unique optimal clustering solution is challenging, as no ground-truth is available in practice. Thus, the selection of the optimal clustering methods (such as $K$-means, density-based OPTICS, or SOM-based clustering) is highly application-dependent. Users may need to apply and combine multiple clustering methods to uncover and identify fault types or fault sub-types in the data effectively.}

% Based on Figure~\ref{fig:results_clustering_over_time}, the proposed framework demonstrates the capability to achieve fault segmentation even with a limited number of samples and online data streaming, which is commonly seen in real-world applications.}

% This indicates that the clustering algorithm is not able to differentiate between damper conditions in the low spring setting, highlighting potential limitations in distinguishing certain fault types. 

% the availability of additional samples results in clearer clusters. These figures provide an intuition of the clustering results and do not represent the actual distances in the feature space.

% For instance, the lSnD samples (Condition \#4), previously considered as outliers, start to form a cluster, also other clusters become more distinct as more samples are added. 

% \begin{figure*}[htbp]
% \centerline{\includegraphics[width=\textwidth]{xai_spectogram_results_4064_mt0_100_mf0_128_n_iter1_channel2.pdf}}
% \caption{(a) - (e) The representative instances for each condition, which have the shortest distance in CAE latent space to each cluster centroid. From top to bottom rows: the vibration signals in the axial direction, their corresponding spectrograms, and the corresponding attribution maps.}
% \label{fig:centroid_samples_axi_mic}
% \end{figure*}
% (f) query sample I with unknown condition (g) query sample II with unknown condition.

\subsection{Fault Diagnostics - Explainable Artificial Intelligence (XAI)-guided Diagnostics}
\label{sec:results_xai}
% After segmenting samples into different clusters, 
% This section evaluates the performance of fault diagnostics based on XAI methods. The qualitative evaluation of the XAI attributions with the additional classifier $C_{\theta_c}(\cdot)$ based on the clustering results are visualized in the last row of Figure~\ref{fig:centroid_samples_axi_mic} for the five representative instances from (a) to (e). 

\begin{figure*}[htbp]
\centerline{\includegraphics[width=\textwidth]{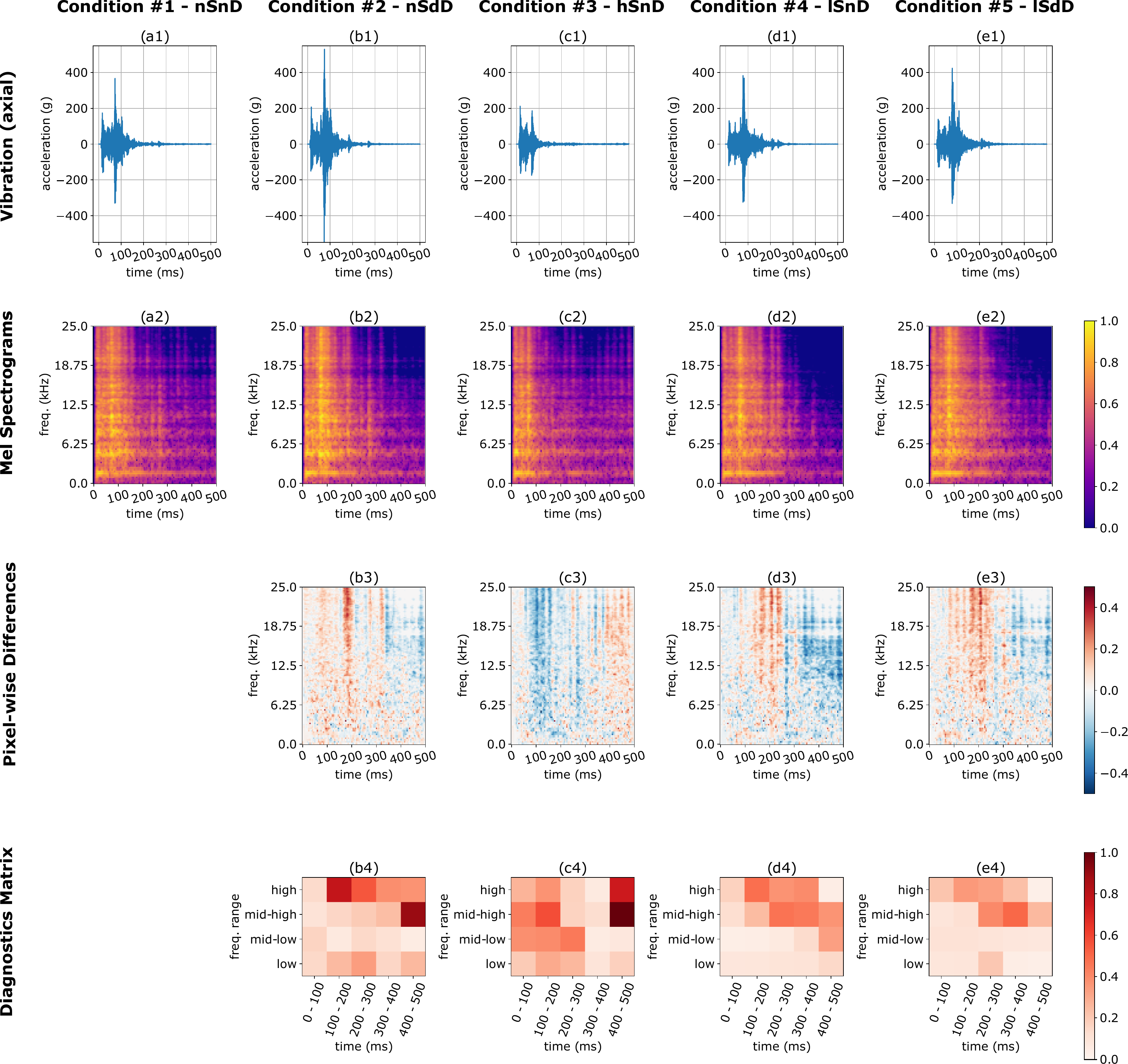}}
\caption{\blue{The representative instances for each condition are selected as the closest samples to each cluster centroid in the CAE latent space. Each column corresponds to different condition: (a) Condition \#1 nSnD, (b) Condition \#2 nSdD, (c) Condition \#3 hSnD, (d) Condition \#4 lSnD, and (e) Condition \#5 lSdD. Each row represents different figures: 1. vibration signals in the axial direction, 2. corresponding Mel spectrograms, 3. corresponding pixel-wise differences between faulty and  healthy spectrograms, 4. diagnostics matrix using max pooling based on attribution maps. Note: No difference spectrogram or diagnostics matrix is provided for the healthy condition (Condition \#1).}}
\label{fig:centroid_samples_axi_mic}
\end{figure*}

\blue{This section evaluates the performance of fault diagnostics using XAI, focusing on the results of $K$-means with $K=5$ for conciseness. First, the cluster centroids in the CAE latent space are used to identify a set of representative instances, which are the data points closest to each centroid. These instances are considered the most representative samples for each condition. The vibration signals of these samples in the axial direction, along with their corresponding spectrograms, are presented in the first and the second rows of Figure~\ref{fig:centroid_samples_axi_mic}, namely in Figure~\ref{fig:centroid_samples_axi_mic} from (a1) to (e1) and from (a2) to (e2).}

\blue{Before applying the XAI methods for fault diagnostics, we introduce an initial analysis step by computing the pixel-wise differences between the healthy and faulty normalized spectrograms, represented as $\bm{x_{f}} - \bm{x_h}$. These differences, shown in Figure~\ref{fig:centroid_samples_axi_mic} (b3), (c3), (d3), and (e3), provide an initial fault visualization. Here, $\bm{x_{f}}$ represents the faulty spectrogram, while $\bm{x_{h}}$ denotes the healthy spectrogram. Red and blue regions indicate areas where the faulty spectrogram exhibits higher or lower values, respectively, compared to the healthy reference.} 

\blue{Under low spring tension conditions (Condition \#4 and \#5), as shown in Figure~\ref{fig:centroid_samples_axi_mic} (d3) and (e3), the spectrogram differences have similar patterns, with higher values around \qty{200}{ms}, particularly in the high frequency regions, and lower values after \qty{300}{ms} compared to the healthy spectrogram. In contrast, under high spring tension (Condition \#3), the increased stiffness reduces vibration amplitude, leading to a shorter vibration duration and faster damping, as seen in the vibration signals in Figure~\ref{fig:centroid_samples_axi_mic} (c1). This is reflected in the spectrogram differences, where lower values appear around \qty{100}{ms}, and the red and blue regions are inverted compared to the low spring tension condition. Similarly, when the damper degrades and its viscosity decreases, the vibration amplitude increases, as shown in Figure~\ref{fig:centroid_samples_axi_mic} (b1) and (e1). In these cases, red regions appear more prominently in the higher frequency areas, indicating a stronger vibration response due to reduced damping efficiency.} 

% In summary, the faults detected in Section~\ref{sec:results_anomaly} can be clustered, even if the model was only trained on healthy data and had never seen faulty samples before, achieving fault segmentation.}
% discussed in Section~\ref{sec:results_clustering}. 
% The first row and the second rows of Figure~\ref{fig:centroid_samples_axi_mic} correspond to the raw vibration signals in axial direction and its corresponding Mel spectrogram. 
%  and two example query samples (f) and (g)
% The first row represents raw time-series vibration signals, while the second row displays  the time-frequency spectrograms attribution maps for vibration in axial direction. 
% \blue{By applying XAI method, The third and the fourth rows in Figure~\ref{fig:centroid_samples_axi_mic}. present Attribution maps max pooling and diagnostics matrix}

\blue{The XAI method, Integrated Gradients, as described in Section~\ref{sec:xai_fault_diagnostics}, is applied to the trained classifier to generate attribution maps and diagnostics matrices, providing interpretability for the cluster assignment obtained from the fault segmentation step. Higher attribution values in these maps highlight the features that contribute most to assigning a sample to a specific cluster. By analyzing these maps, domain experts can identify  potential fault types by recognizing similarities and differences across conditions. The diagnostics matrix for each faulty condition is presented in Figure~\ref{fig:centroid_samples_axi_mic} (b4), (c4), (d4), and (e4).} 

% \blue{In this section, the average spectrograms of healthy samples are used as the baseline for gradient calculations. However, due to the sparsity of the attribution maps, interpreting the raw maps remains challenging, even for domain experts. Therefore, max pooling operations with a pooling size of (32, 20) are applied to refine the maps and generate a "diagnostics matrix", as shown in the last row of Figure~\ref{fig:centroid_samples_axi_mic}. The diagnostics matrix in this case has a temporal resolution of five intervals (\qty{100}{ms} each) and a frequency resolution divided into four bands (low, mid-low, mid-high, high). Each element in the matrix represents the attribution value for a specific time-frequency region, facilitating a more interpretable process for fault diagnostics.} 

\blue{The diagnostics matrices in Figure~\ref{fig:centroid_samples_axi_mic} (d4) and (e4) suggest that these two samples correspond to similar fault types, as indicated by their similar diagnostics matrices. High attribution values are observed between \qty{100}{ms} and \qty{300}{ms} in the high-frequency regions and between \qty{200}{ms} and \qty{400}{ms} in the mid-high-frequency regions, highlighting shared fault characteristics. In Figure~\ref{fig:centroid_samples_axi_mic} (b4), the presence of a red region around \qty{200}{ms} suggests an additional vibration event occurring across all frequency ranges. The diagnostics matrix further reinforces this observation, as the highlighted high-frequency region between \qty{100}{ms} and \qty{300}{ms} indicates that this additional event plays a key role in cluster assignment during the fault segmentation step. These diagnostics matrices enhance the interpretability of the fault segmentation results obtained from unsupervised clustering methods, providing valuable insights into the distinguishing features of different fault conditions.}

% Similarly, the sample in Figure~\ref{fig:centroid_samples_axi_mic} (c) shows a distinct pattern and could represent a different fault type, with the high-frequency region around \qty{400}{ms} to \qty{500}{ms} being highlighted. 

% \blue{An example of using the proposed framework would be: a new vibration sample is collected and detected as faulty in the fault detection step using CAE. This sample is clustered into yellow cluster in Figure~\ref{fig:results_clustering_tsne}. By calculating the difference to the healthy sample and applying IG, the third and the four row of Figure~\ref{fig:centroid_samples_axi_mic} (b) can be obtained. To perform fault diagnostics, first the red region around \qty{200}{ms} is observed, meaning that an additional vibration event across all frequency range is present in the signal. From the diagnostics matrix, the highlighted area in high-frequency between \qty{100}{ms} and \qty{300}{ms} further confirm that these additional event is causing the cluster assignment. With these diagnostics matrix, interpretation of the fault segmentation results with unsupervised clustering methods is possible.}

\begin{figure}[htbp]
\centerline{\includegraphics[width=100mm]{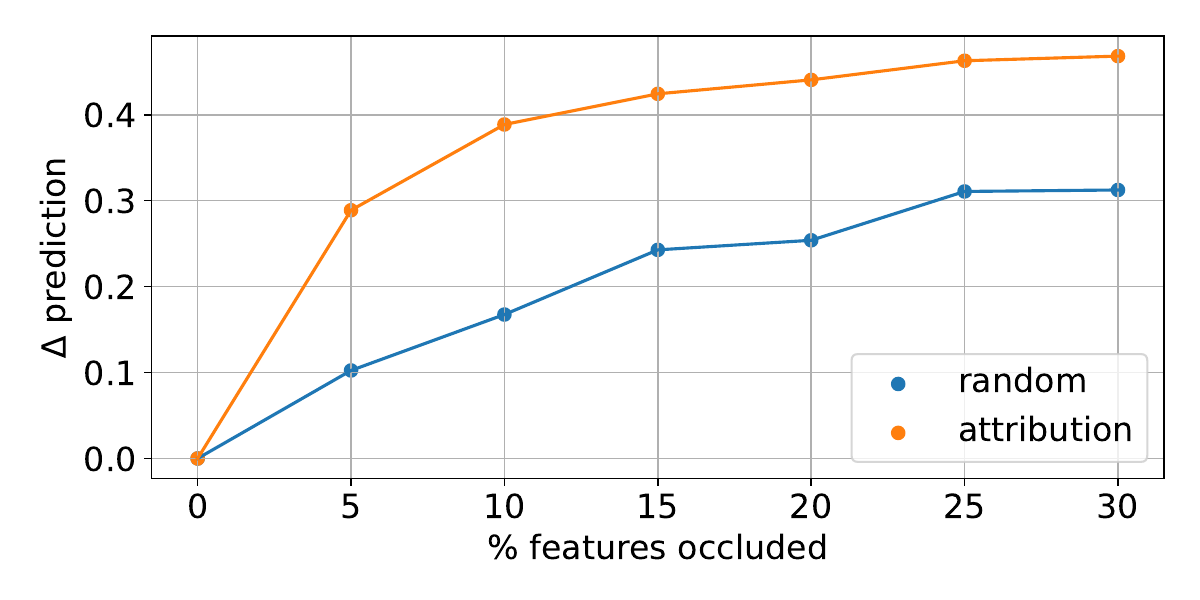}}
\caption{Faithfulness of the explanations evaluated by representing the change in classifier prediction confidence ($\Delta$) as a function of the percentage of features (pixels in spectrogram) occluded, averaged over the entire dataset. The impact is higher for attribution-based occlusion than for random occlusion, showing that the features with higher attribution values are indeed important.}
\label{fig:faithfulness_perturb}
\end{figure}

To further evaluate the faithfulness of the XAI attribution maps, the changes in the classifier's prediction confidence are represented in Figure~\ref{fig:faithfulness_perturb} for both random occlusion and attribution-based occlusion, as described in Section~\ref{sec:evaluation}. We gradually occlude between $0\%$ (original spectrogram) and $30\%$ of the total input features in the spectrograms by replacing them with zeros. For random occlusion, features are occluded randomly, while for attribution-based occlusion, features are selected in descending order of their attribution values. In both cases, the prediction delta $\Delta$ increases with the percentage of occluded features, but the curve is significantly higher for the attribution-based occlusion. The changes in predictions are higher for attribution-based occlusion than for random occlusion, confirming that the features with higher attribution values identified by the XAI method are indeed important for the assignment to a specific cluster.

\subsection{Contribution of each Sensor to Fault Segmentation}

% ~\cite{yang2019chaotic} Sensitivity of Accelerometer’s Position
In this section, we examine the significance of each vibration direction and microphone in distinguishing between different faults when performing fault segmentation using $K$-means. We report four evaluation metrics in Table~\ref{tab:ablation_study_sensor}: ARI score, homogeneity score $h$, completeness score $c$, and v-measure $v$, as described in Section~\ref{sec:evaluation}. 

% The name ablation study in this section is not related to the removal of any component in the proposed network but rather using different input sensors or input time or bandwidth frequency ranges.

% Statistical features as baseline for comparison 

% To understand how long should the recorded length be and what frequency range should be recorded, the time-frequency maps are masked. In this case all four sensor recordings are used, namely, the setting A in Table~\ref{table:ablation}. The time is divided into five cases, from 0 to 100, 200, 300, 400, and 500ms. Similarly, the frequency is divided in to four cases, from 0 to 6.25, 12.5, 18.75, 25kHz. The results of evaluation metrics are shown in Figure xx.

% To know which sensors or directions are important for clustering, the results from the ablation study as defined in Table~\ref{table:ablation} are shown in Figure xx. 
Using signals recorded from all three accelerometers and one microphone achieved the highest scores among all settings, with all four evaluation metrics exceeding 0.9, indicating well-separated clusters. In contrast, the worst performance is observed using only the microphone signals. Comparable performances are achieved using only one accelerometer. This discrepancy may stem from the fact that the microphone is not directly mounted on the CB structure, leading to a loss of information during the transmission of vibrations through the air. Unlike the accelerometers, which are directly mounted on the CB, the microphone signal is more prone to coupling with environmental noise. The direction of the accelerometer installation does not show a significant difference based on this experimental dataset. However, installation in the vertical and axial directions performs better than in the horizontal directions. 

In summary, to best distinguish between the fault types, all four sensors, including three accelerometers and one microphone, should ideally be used. However, the sensors installed in different directions contain highly redundant information. Using only the vertical or axial accelerometers yields comparable clustering performance for distinguishing faulty samples. Considering installation costs and practicality, installing an accelerometer in the axial direction is the preferred option for this experimental setup. Alternatively, a single vibration sensor capable of measuring three-directional vibrations could also be considered.

\begin{table}[h]
    \centering
        \caption{Influence study on sensors. hor: accelerometer in horizontal direction, ver: accelerometer in vertical direction, axi: accelerometer in axial direction, mic: microphone. The symbol $\uparrow$ means the higher the value is, the better separated the clusters. The best score is 1, where clusters are well separated.}
    \begin{tabular}{|c|c|c|c|c|}
    
    \hline
        Sensor(s) & ARI $\uparrow$ & $h$ $\uparrow$ & $c$ $\uparrow$ & $v$ $\uparrow$\\
    \hline
         hor, ver, axi, mic & \textbf{0.9045} & \textbf{0.9013} & \textbf{0.9018} & \textbf{0.9015}\\
    \hline
        hor & 0.8287 & 0.8237 & 0.8245 & 0.8241\\
    \hline
        ver & 0.8607 & 0.8587 & 0.8651 & 0.8619\\
    \hline
        axi & 0.8673 & 0.8737 & 0.8811 & 0.8774\\
    \hline
        mic & 0.7839 & 0.8018 & 0.8063 & 0.8040\\
    \hline
    \end{tabular}

    \label{tab:ablation_study_sensor}
\end{table}

%% file: conclusions.tex
\section{Conclusions}
\label{sec:conclusions}

In this study, we propose an unsupervised fault detection and segmentation framework for condition monitoring of CBs with an XAI approach integrated into the framework to achieve fault diagnostics and assist domain experts in identifying potential high-voltage CB fault types. The effectiveness of the proposed framework was validated on a mechanical switching dataset collected in the laboratory with different fault types. The clustering results using three different clustering methods have demonstrated the framework's flexibility and feasibility in grouping healthy and unknown faulty samples into distinct clusters. Furthermore, the results from XAI further explain the clustered samples, achieving fault diagnostics even if the fault type has not yet been observed and ground-truth labels are not available during training.

% Furthermore, the proposed unsupervised fault detection and segmentation framework for condition monitoring of CBs was evaluated on a mechanical switching dataset collected in the laboratory. The clustering results have demonstrated the framework's feasibility in grouping healthy and unknown faulty samples into different clusters. 

% The results from XAI were consistent with the ablation study, highlighting that the majority of attributions are contained within the first \qty{300}{ms} and in the frequency range below \qty{12.5}{kHz} . 
% , with all four evaluation metrics (ARI, $h$, $c$, $v$ scores) exceeding 0.9. 

% In addition, an ablation study was conducted to understand
% In addition, the impact of different directions of accelerometers, optimal recording duration, and frequency range are investigated. The accelerometers contain more information than the microphone, and both vertical and axial directions have similar performance if only one sensor is used. The optimal recording length is \qty{300}{ms} or \qty{400}{ms}, resulting in all four evaluation metrics around 0.8 and 0.9, respectively. Regarding the recording bandwidth frequency range, an upper limit of \qty{12.5}{kHz} was deemed possible, as the performance saturated beyond this point. 
% However, for optimal performance, all four sensors, a full recording length of 500 ms, and a frequency range up to 25 kHz should be used. 

% Depending on what the aim is, if one only wants to perform anomaly detection but not further clustering and explanation, using only microphone is sufficient. 

This work highlights future research directions such as distinguishing between different severity of the same fault type and understanding how different levels of severity evolve in the clustering space. Finally, the transferability of the proposed framework to different CBs of the same type, to CBs of the same operating mechanism but different manufacturers, or even to different CB types is left for future research.

% The next step is also to apply this approach to CB data collected with current interruption, either in laboratories or in substations. These scenarios could have more noise due to electromagnetic interference (EMI). 
% It is worth mentioning that microphone is much simpler to install than accelerometer as it does not need to be attached to the structure. 

% \todo[inline]{The framework is not specific to k-means clustering methods, it could be any clustering method (but in this case, k-means is suited. Noise, imbalacne -> GMM, ...)}

\section*{Acknowledgement}
This work is part of a project that is financially supported by the Swiss Federal Office of Energy, research program \textbf{energy research and cleantech}.